%% file: AdSpark.tex
\documentclass{article} 
\usepackage{iclr2027_conference,times}

\input{math_commands.tex}

\usepackage{hyperref}
\usepackage{url}
\usepackage{algorithm}
\usepackage{algorithmic}
\usepackage{booktabs}
\usepackage{amsmath}
\usepackage{multirow}
\usepackage[most]{tcolorbox}
\usepackage{xcolor}
\usepackage{listings}
\usepackage{caption}

\usepackage{newfloat}
\usepackage{listings}
\DeclareCaptionStyle{ruled}{labelfont=normalfont,labelsep=colon,strut=off} 
\floatstyle{ruled}
\newfloat{listing}{tb}{lst}{}
\floatname{listing}{Listing}

\usepackage{amssymb}
\usepackage{pifont}
\usepackage{colortbl}
\usepackage{enumitem}
\usepackage{pifont}
\usepackage{makecell}

\definecolor{mycolor_blue}{HTML}{E7EFFA}
\definecolor{mycolor_green}{HTML}{E6F8E0}
\definecolor{mycolor_gray}{HTML}{ECECEC}
\definecolor{pearDark}{HTML}{2980B9}
\definecolor{citecolor}{HTML}{2980b9}
\definecolor{linkcolor}{HTML}{c0392b}
\definecolor{blank}{HTML}{000000}
\definecolor{sem}{HTML}{2E75B6}
\definecolor{tok}{HTML}{F3B000}

\usepackage{pifont}     
\definecolor{my_green}{RGB}{51,102,0}
\definecolor{my_red}{RGB}{204, 0, 0}
\renewcommand{\checkmark}{\textcolor{my_green}{\ding{51}}} 
\newcommand{\crossmark}{\textcolor{my_red}{\ding{55}}} 
\definecolor{paired-light-blue}{RGB}{198, 219, 239}
\definecolor{paired-dark-blue}{RGB}{49, 130, 188}
\definecolor{paired-light-orange}{RGB}{251, 208, 162}
\definecolor{paired-dark-orange}{RGB}{230, 85, 12}
\definecolor{paired-light-green}{RGB}{199, 233, 193}
\definecolor{paired-dark-green}{RGB}{49, 163, 83}
\definecolor{paired-light-purple}{RGB}{218, 218, 235}
\definecolor{paired-dark-purple}{RGB}{117, 107, 176}
\definecolor{paired-light-gray}{RGB}{217, 217, 217}
\definecolor{paired-dark-gray}{RGB}{99, 99, 99}
\definecolor{paired-light-pink}{RGB}{222, 158, 214}
\definecolor{paired-dark-pink}{RGB}{123, 65, 115}
\definecolor{paired-light-red}{RGB}{231, 150, 156}
\definecolor{paired-dark-red}{RGB}{131, 60, 56}
\definecolor{paired-light-yellow}{RGB}{231, 204, 149}
\definecolor{paired-dark-yellow}{RGB}{141, 109, 49}  
\definecolor{myblue}{RGB}{218,232,252}
\definecolor{mygray}{RGB}{220,220,220}
\definecolor{mypink}{RGB}{251,49,153}

\title{AdSpark: A Large-Scale Dataset and Benchmark for Product-Centric Advertisement Video Generation}

\author{%
\begin{tabular}{@{}l@{}}
Zhifei Yang$^{1}$,
Zhao Jiang$^{2}$,
Keyang Lu$^{1}$,
Honghe Zhu$^{2}$,
Zheng Zhang$^{2,*}$,
Jingjing Lv$^{2}$,\\
Changping Peng$^{2}$,
Ching Law$^{2}$,
Zhen Xiao$^{1,*}$
\end{tabular}\\[3pt]
\normalfont
$^{1}$Peking University
\quad
$^{2}$JD.com
\quad
$^{*}$Corresponding authors
}

\iclrfinalcopy 
\begin{document}

\maketitle
\lhead{\small AdSpark: A Large-Scale Dataset and Benchmark for Product-Centric Advertisement Video Generation}

\input{Sections/0_abstract}

\input{Sections/1_introduction}

\input{Sections/2_related_work}

\input{Sections/3_method}

\input{Sections/4_experiments}

\input{Sections/5_conclusion}

\bibliography{iclr2027_conference}
\bibliographystyle{iclr2027_conference}

\appendix
\input{Sections/X_supplementary}

\end{document}

%% file: math_commands.tex
\usepackage{amsmath,amsfonts,bm}

\def\eqref#1{equation~\ref{#1}}

\def\1{\bm{1}}

\DeclareMathAlphabet{\mathsfit}{\encodingdefault}{\sfdefault}{m}{sl}
\SetMathAlphabet{\mathsfit}{bold}{\encodingdefault}{\sfdefault}{bx}{n}



%% file: Sections/0_abstract.tex
\begin{abstract}
Product-centric advertisement video generation aims to create promotional videos that preserve fine-grained product identity while presenting selling points through coherent multi-shot narratives. However, this emerging task remains underexplored due to the lack of large-scale advertisement-specific datasets and comprehensive evaluation frameworks.
To address this gap, we introduce \textbf{AdSpark}, a large-scale dataset and benchmark for product-centric advertisement video generation, based on data from a major e-commerce platform. \textit{AdSpark-300K} contains approximately 300K reference image--prompt--video triplets, comprising a real-world subset and a synthetic subset.
Each sample provides structured advertisement annotations, including product identity annotations, selling-point descriptions, creative plans, and aligned audio scripts, enabling models to learn product preservation and advertisement-oriented visual storytelling.
We further propose \textit{AdSpark-Bench}, a diagnostic benchmark that evaluates generated advertisements across six dimensions, including visual quality, product fidelity, instruction adherence, temporal coherence, audio alignment, and advertisement effectiveness.
Based on AdSpark-Bench, we evaluate representative models, revealing key challenges in product preservation, multi-shot storytelling, and selling-point visualization. 
Experiments with AdSpark-300K-finetuned models further validate the effectiveness of our dataset. AdSpark provides a unified dataset and benchmark for future research, and we will release the dataset upon acceptance.
\end{abstract}

%% file: Sections/1_introduction.tex
\section{Introduction}
\input{tex/teaser}

With the rapid growth of e-commerce and short-form video platforms, advertisement videos have become increasingly important for product promotion and  digital marketing. However, traditional advertisement production requires professional designing and filming, making it labor-intensive and costly. 
Recent advances in generative visual modeling, spanning reference-to-video generation~\cite{liu2025phantom, chen2026vino, wang2026refalign} and 3D generation~\cite{yang2025mmgdreamer,yang2026flowscene,wang2025spatial,lu2026yo,wang2025mari,lu2026cadforgeagenticsingleviewcad}, have enabled increasingly realistic, controllable, and scalable synthesis from visual or textual conditions, opening new opportunities for automated advertisement production.

Nevertheless, existing video generation models~\cite{song2026mv, zhang2025kaleido,yuan2026video} are primarily developed for general-purpose visual content generation, with an emphasis on visual quality, semantic alignment, and temporal coherence. In contrast, product-centric advertisement videos additionally require faithful preservation of fine-grained product identity and effective visualization of selling points through coherent multi-shot narratives. Achieving these goals involves coordinated control over scene composition, camera motion, shot transitions, and audio to create a compelling advertising experience. 
Thus, current models~\cite{fang2023alignment,fang2023mask,fanrethinking,song2026mv,hacohen2026ltx} still struggle with \textbf{Product-Centric Advertisement Video Generation~(PC-AVG)}, leaving the task challenging and largely underexplored.

\input{Tables/compare_dataset}

\input{Tables/compare_benchmark}
A major bottleneck is the lack of large-scale, high-quality datasets for PC-AVG. As summarized in Tab.~\ref{table_dataset_comparison}, existing video datasets are primarily developed for open-domain text-to-video generation~\cite{InternVid, Panda-70M}, image-to-video generation~\cite{wu2026consid}, or subject-to-video generation~\cite{yuan2026opens2v, zhang2026muss}. However, these datasets mainly consist of generic video--text pairs or reference-conditioned videos and are not specifically curated for advertising scenarios. Moreover, they rarely provide advertisement-specific annotations, such as product identity, selling points, and creative plans, limiting their effectiveness for adapting video generation models to PC-AVG.

Beyond training data, existing benchmarks are insufficient for evaluating PC-AVG. As shown in Tab.~\ref{tab:benchmark_comparison}, current benchmarks~\cite{vbench++, yuan2026opens2v,yang2026vidtext} mainly focus on general video generation capabilities, including visual quality, semantic alignment, and temporal coherence. However, these assessment frameworks overlook key advertisement-specific aspects, such as fine-grained product preservation, selling-point visualization, and advertisement effectiveness. This gap makes it difficult to evaluate whether generated videos not only achieve high perceptual quality, but also follow the intended advertisement design and effectively communicate commercial messages. Therefore, existing benchmarks cannot fully characterize the capabilities required for PC-AVG, highlighting the need for a dedicated benchmark.

To address these limitations, we introduce AdSpark, a large-scale dataset and benchmark for PC-AVG. As shown in Fig.~\ref{fig:teaser}, AdSpark-300K contains 300K high-quality reference image-prompt-video triplets, including 100K real-world and 200K synthetic advertisement videos. It covers 40 diverse product categories, over 3,000 sub-categories, and 651 hours of video. Each sample provides structured annotations for product identity, selling points, creative plans, and aligned audio scripts, enabling faithful product preservation and advertisement-oriented visual storytelling.
Alongside the dataset, we design AdSpark-Bench, a diagnostic benchmark for PC-AVG that evaluates the capabilities of video generation models in advertisement scenarios. Beyond conventional video quality assessment, AdSpark-Bench analyzes whether models can preserve product identity, follow creative plans, present selling points, and generate compelling and commercially effective advertisements. 

Based on AdSpark-300K, we finetune a reference-to-video generation model~\cite{hacohen2026ltx} and conduct comprehensive evaluations of open-source and proprietary models on AdSpark-Bench. The results validate the effectiveness of AdSpark-300K and provide insights into the capabilities and challenges of current models for PC-AVG. Our main contributions are summarized as follows:
\begin{itemize}
    \item We introduce \textbf{AdSpark-300K}, the first large-scale dataset for PC-AVG, containing 300K high-quality reference image-prompt-video triplets with structured annotations for product identity, selling points, creative plans, and audio scripts.

    \item We propose \textbf{AdSpark-Bench}, a comprehensive and diagnostic benchmark that extends conventional video evaluation with advertisement-specific criteria, including product fidelity, shot compliance, selling-point realization, and advertisement effectiveness.

    \item We conduct comprehensive evaluations on AdSpark-Bench, revealing the limitations of existing models for PC-AVG and demonstrating that fine-tuning a representative video generation model on AdSpark-300K substantially improves advertisement-oriented generation. 

\end{itemize}

%% file: tex/teaser.tex
\begin{figure*}
    \centering
    \includegraphics[width=\linewidth]{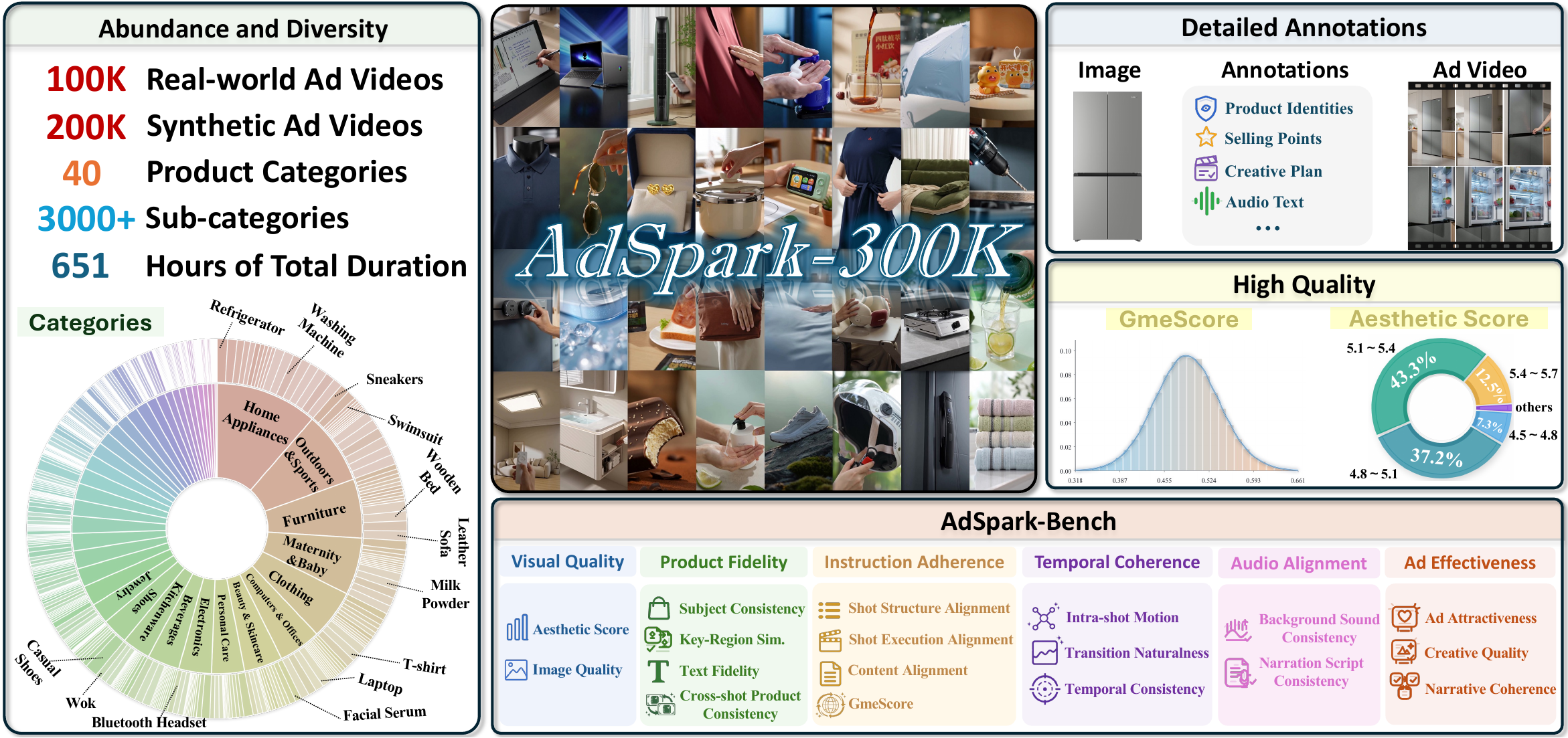}
   \caption{\textbf{Overview of AdSpark-300K and AdSpark-Bench.} AdSpark-300K is a large-scale, high-quality product-centric advertisement video dataset spanning real-world and synthetic advertisements, with structured annotations across multiple dimensions. AdSpark-Bench provides a comprehensive evaluation of general video quality and advertisement-specific capabilities.}
    \label{fig:teaser}
\end{figure*}

%% file: Tables/compare_dataset.tex
\begin{table*}[t]
    \centering
    \caption{
    \textbf{Comparison with existing video generation datasets.}
    Most existing datasets focus on generic text-to-video generation~(T2V), subject-to-video generation~(S2V), or image-to-video generation~(I2V), while lacking advertisement-specific annotations.
    In contrast, AdSpark-300K targets PC-AVG, providing structured advertisement annotations, including product identity, selling points, creative plans (style, scene, and shot design), and audio scripts.
    }
    \label{table_dataset_comparison}
    \resizebox{\textwidth}{!}{
        \begin{tabular}{lcccccccccc}
            \toprule
            \textbf{Dataset}
            & \textbf{Domain}
            & \textbf{Task}
            & \makecell{\textbf{Multi-}\\\textbf{shot}}
            & \makecell{\textbf{Creative}\\\textbf{Plans}}
            & \makecell{\textbf{Selling}\\\textbf{Points}}
            & \makecell{\textbf{Identity}\\\textbf{Annotation}}
            & \makecell{\textbf{Audio}\\\textbf{Script}}
            & \textbf{Clips}
            & \textbf{Resolution}
            & \makecell{\textbf{Average}\\\textbf{Length(s)}} \\
            \midrule

            MSRVTT~\cite{MSR-VTT}
            & Open
            & T2V
            & \crossmark
            & \crossmark
            & \crossmark
            & \crossmark
            & \crossmark
            & 10K
            & 240P
            & 14.4 \\

            WebVid-10M~\cite{WebVid}
            & Open
            & T2V
            & \crossmark
            & \crossmark
            & \crossmark
            & \crossmark
            & \crossmark
            & 10M
            & 360P
            & 18.7 \\

            HD-VG-130M~\cite{HD-130M}
            & Open
            & T2V
            & \crossmark
            & \crossmark
            & \crossmark
            & \crossmark
            & \crossmark
            & 130M
            & 720P
            & 4.9 \\

            Panda-70M~\cite{Panda-70M}
            & Open
            & T2V
            & \crossmark
            & \crossmark
            & \crossmark
            & \crossmark
            & \crossmark
            & 70M
            & 720P
            & 8.6 \\

            InternVid~\cite{InternVid}
            & Open
            & T2V
            & \crossmark
            & \crossmark
            & \crossmark
            & \crossmark
            & \crossmark
            & 234M
            & 720P
            & 11.7 \\

            OpenHumanVid~\cite{openhumanvid}
            & Human
            & T2V
            & \crossmark
            & \crossmark
            & \crossmark
            & \crossmark
            & \crossmark
            & 52.3M
            & 720P
            & 4.9 \\

            Cine250K~\cite{wu2025cinetrans}
            & Movies
            & T2V
            & \checkmark
            & \crossmark
            & \crossmark
            & \crossmark
            & \crossmark
            & 250K
            & 720P
            & 10.7 \\

            OpenS2V-5M~\cite{yuan2026opens2v}
            & Subject
            & S2V
            & \crossmark
            & \crossmark
            & \crossmark
            & \crossmark
            & \crossmark
            & 5.4M
            & 720P
            & 6.6 \\

            MuSS~\cite{zhang2026muss}
            & Movies
            & S2V
            & \checkmark
            & \crossmark
            & \crossmark
            & \crossmark
            & \crossmark
            & 700K
            & 720P
            & 5.1 \\

            ConsIDVid~\cite{wu2026consid}
            & Rigid Objects
            & I2V
            & \crossmark
            & \crossmark
            & \crossmark
            & \crossmark
            & \crossmark
            & 44.3K
            & 480P
            & 8.4 \\

            \midrule

            \rowcolor{myblue}
            \textbf{AdSpark-300K}
            & \textbf{Product Ads}
            & \textbf{PC-AVG}
            & \checkmark
            & \checkmark
            & \checkmark
            & \checkmark
            & \checkmark
            & \textbf{300K}
            & \textbf{720P}
            & \textbf{7.8} \\

            \bottomrule
        \end{tabular}
    }
\end{table*}

%% file: Tables/compare_benchmark.tex
\begin{table*}[t]
    \centering
    \caption{
    \textbf{Comparison of AdSpark-Bench with existing video generation benchmarks.}
    Existing benchmarks focus on general video quality, while AdSpark-Bench further evaluates product fidelity, selling-point realization, and advertisement effectiveness.
     \underline{\checkmark} indicates that the corresponding dimension is evaluated, but less comprehensively than in AdSpark-Bench. 
    }
    \label{tab:benchmark_comparison}
    \resizebox{\textwidth}{!}{
        \begin{tabular}{lcccccccc}
        \toprule
        \textbf{Benchmark}
        & \makecell{\textbf{Visual}\\\textbf{Quality}}
        & \makecell{\textbf{Script}\\\textbf{Adherence}}
        & \makecell{\textbf{Temporal}\\\textbf{Coherence}}
        & \makecell{\textbf{Product}\\\textbf{Fidelity}}
        & \makecell{\textbf{Shot}\\\textbf{Compliance}}
        & \makecell{\textbf{Audio}\\\textbf{Alignment}}
        & \makecell{\textbf{Selling-point}\\\textbf{Realization}}
        & \makecell{\textbf{Advertisement}\\\textbf{Effectiveness}} \\
        \midrule

        Make-a-Video-Eval \cite{Make-a-video}
        & \checkmark
        & \checkmark
        & \crossmark
        & \crossmark
        & \crossmark
        & \crossmark
        & \crossmark
        & \crossmark \\

        FETV \cite{FETV}
        & \checkmark
        & \checkmark
        & \checkmark
        & \crossmark
        & \crossmark
        & \crossmark
        & \crossmark
        & \crossmark \\

        T2VScore \cite{T2VScore}
        & \checkmark
        & \checkmark
        & \checkmark
        & \crossmark
        & \crossmark
        & \crossmark
        & \crossmark
        & \crossmark \\

        EvalCrafter \cite{evalcrafter}
        & \checkmark
        & \checkmark
        & \checkmark
        & \crossmark
        & \crossmark
        & \crossmark
        & \crossmark
        & \crossmark \\

        VBench \cite{Vbench}
        & \checkmark
        & \checkmark
        & \checkmark
        & \crossmark
        & \crossmark
        & \crossmark
        & \crossmark
        & \crossmark \\

        VBench++ \cite{vbench++}
        & \checkmark
        & \checkmark
        & \checkmark
        & \crossmark
        & \crossmark
        & \crossmark
        & \crossmark
        & \crossmark \\

        ChronoMagic-Bench \cite{chronomagicbench}
        & \checkmark
        & \checkmark
        & \checkmark
        & \crossmark
        & \crossmark
        & \crossmark
        & \crossmark
        & \crossmark \\

        ConsisID-Bench \cite{consisid}
        & \checkmark
        & \checkmark
        & \checkmark
        & \underline{\checkmark}
        & \crossmark
        & \crossmark
        & \crossmark
        & \crossmark \\

        Alchemist-Bench \cite{msrvttpersonlize}
        & \checkmark
        & \checkmark
        & \checkmark
        & \underline{\checkmark}
        & \crossmark
        & \crossmark
        & \crossmark
        & \crossmark \\

        A2 Bench \cite{skyreelsa2}
        & \checkmark
        & \checkmark
        & \checkmark
        & \underline{\checkmark}
        & \crossmark
        & \crossmark
        & \crossmark
        & \crossmark \\

        OpenS2V-Eval \cite{yuan2026opens2v}
        & \checkmark
        & \checkmark
        & \checkmark
        & \underline{\checkmark}
        & \crossmark
        & \crossmark
        & \crossmark
        & \crossmark \\

        VACE-Bench \cite{jiang2025vace}
        & \checkmark
        & \checkmark
        & \checkmark
        & \underline{\checkmark}
        & \underline{\checkmark}
        & \crossmark
        & \crossmark
        & \crossmark \\

        MultiShotMaster \cite{wang2026multishotmaster}
        & \checkmark
        & \checkmark
        & \checkmark
        & \underline{\checkmark}
        & \underline{\checkmark}
        & \crossmark
        & \crossmark
        & \crossmark \\

        MSAVBench \cite{wei2026msavbench}
        & \checkmark
        & \checkmark
        & \checkmark
        & \underline{\checkmark}
        & \underline{\checkmark}
        & \checkmark
        & \crossmark
        & \crossmark \\

        \midrule

        \rowcolor{myblue}
        \textbf{AdSpark-Bench}
        & \checkmark
        & \checkmark
        & \checkmark
        & \checkmark
        & \checkmark
        & \checkmark
        & \checkmark
        & \checkmark \\

        \bottomrule
        \end{tabular}
    }
\end{table*}

%% file: Sections/2_related_work.tex
\section{Related Work}
\paragraph{Datasets for Advertisement Video Generation.}
Large-scale video-text datasets, such as WebVid-10M~\cite{WebVid}, Panda-70M~\cite{Panda-70M}, and InternVid~\cite{InternVid}, have advanced open-domain video generation. However, they are not curated for advertising scenarios and lack product-specific annotations. 
Recent multi-shot video datasets, such as Cine250K~\cite{wu2025cinetrans} and MuSS~\cite{zhang2026muss}, further support coherent generation across multiple shots by providing shot-level video structures and cross-shot supervision, but mainly focus on general cinematic or narrative content. 
Reference-based video generation datasets, including human-centric~\cite{openhumanvid,animateanyone} and subject-consistent~\cite{consisid,wu2026consid,yuan2026opens2v} datasets, improve subject consistency by preserving referenced subjects but overlook fine-grained product identity and commercial generation requirements. In contrast, AdSpark-300K targets product-centric advertisement video generation with structured annotations covering product identities, selling points, creative plans, and audio scripts.

\paragraph{Benchmarks for Advertisement Video Generation.}
Video generation benchmarks have progressed from perceptual quality evaluation to comprehensive assessment of generation capabilities. Representative benchmarks, including VBench~\cite{Vbench}, VBench++~\cite{vbench++}, EvalCrafter~\cite{evalcrafter}, ChronoMagic-Bench~\cite{chronomagicbench}, and FETV~\cite{FETV}, evaluate videos from multiple perspectives, such as visual quality, semantic alignment, and temporal coherence. 
Recent reference-conditioned benchmarks~\cite{consisid,skyreelsa2,yuan2026opens2v} further incorporate subject consistency, while multi-shot benchmarks additionally evaluate cross-shot consistency and shot-level controllability~\cite{wang2026multishotmaster}.
However, these benchmarks mainly target general video generation and overlook advertisement-specific demands such as product fidelity, selling-point realization and advertisement effectiveness. AdSpark-Bench fills this gap by providing a dedicated evaluation framework for PC-AVG.

%% file: Sections/3_method.tex
\section{AdSpark-300K}
\input{tex/pipeline}
As illustrated in Fig.~\ref{fig:data_pipeline}, AdSpark-300K comprises complementary real-world and synthetic subsets, each developed through a rigorous multi-stage pipeline of filtering, annotation, and quality control. Built upon real-world product assets and associated metadata collected from an e-commerce platform with proper authorization, both subsets share a unified data format containing a reference image, a product-centric advertisement video, and structured annotations.
In this section, we will present the construction of each subset and summarize the overall dataset statistics.

\subsection{Real-World Advertisement Curation}
\paragraph{Multi-stage Filtering.} 
We collect paired advertisement videos, product reference images, and foreground masks from a major e-commerce platform. Given these paired assets, we first use SAM2~\cite{sam2} to segment and track the target product throughout the video, as show in Fig.~\ref{fig:data_pipeline}(a). We then apply a two-stage filtering procedure to retain reliable product-centric advertisement clips. 
In the first stage, frame-level geometric and temporal filtering removes frames where the product is too small, the mask area changes abruptly, the centroid shifts substantially, or the mask touches the image boundary, as these cases indicate poor visibility or unstable tracking.
In the second stage, we group consecutive valid frames into 5--20 second candidate clips and perform VLM-based quality filtering using Qwen3-VL-235B~\cite{bai2025qwen3}, guided by the product reference image, to evaluate product identity consistency, product visibility, and overlay artifacts. Clips failing any criterion are discarded.

\paragraph{Shot Segmentation and Automatic Annotation.}
After obtaining the accepted temporally continuous clips, we use TransNetV2~\cite{soucek2024transnet} to segment them into shots. Since the original videos lack structured advertisement annotations, Qwen3-VL-235B~\cite{bai2025qwen3} is used to automatically annotate product identities, selling points, and creative plans (scene, style, and shot design) from representative frames. Whisper-Large-v3-Turbo~\cite{radford2023robust} is used to transcribe the audio and align narration text with the detected shots. Finally, human annotators verify the videos and annotations, correcting errors and removing low-quality or unreliable samples.
Only the videos that pass this final manual inspection are included in AdSpark-300K.

\subsection{Synthetic Advertisement Construction}
\paragraph{SKU Pool Preparation.}
Given the scarcity of high-quality real-world advertisement videos, we construct a synthetic subset complementary to the real-world subset to enhance the diversity and coverage of AdSpark-300K. Starting from products with reference images and associated attributes, we curate a high-quality stock-keeping unit~(SKU) pool through three-stage quality, suitability, and category filtering. Specifically, we exclude SKUs with low-quality reference images, unsuitable categories, cluttered backgrounds, multiple dominant objects, or severe occlusion to ensure reliable product-centric generation.

\paragraph{Structured Advertisement Planning.} 
For each retained product, we employ GPT-5.5 to generate structured advertisement annotations conditioned on the reference image and product attributes. 
The annotations serve as an intermediate representation for video generation, comprising product identity, selling-point descriptions, creative plans covering scene, style, and shot design, and aligned audio scripts.
In particular, selling points are translated into visualizable actions and effects, enabling the generated videos to demonstrate product functions and commercial appeals rather than merely describe them. The shot-level designs further specify temporal organization, camera motion, and shot type for multi-shot generation. To improve annotations reliability, we employ Gemini-3.1-Pro-Preview as an independent reviewer in an iterative review-and-refinement process. 
Given the reference image and generated annotations, it evaluates whether the annotations accurately describe the product identity and visual details, align with the selling points, and specify coherent shot designs.
Annotations that fail these criteria are revised by GPT-5.5 according to the reviewer feedback.

\paragraph{Video Generation.}
The structured annotations are used to synthesize advertisement videos with multiple state-of-the-art models, including Seedance 2.0~\cite{seedance2026seedance}, Happy Horse, and Kling~\cite{team2025kling}, improving generation diversity and reducing model-specific bias. We further conduct human inspection to assess overall video quality, including product fidelity, visual plausibility, motion quality, selling-point realization, and audio alignment, discarding videos with identity drift, severe artifacts, ineffective selling-point presentation, or audio-visual mismatch.

\paragraph{Unified Data Representation.}
After both construction pipelines, all samples share the same representation. Each sample contains a product reference image, an advertisement video, and structured advertisement annotations comprising product identity, selling-point descriptions, creative plans covering scene, style, and shot design, and aligned audio scripts. 
This unified annotation format enables AdSpark-300K to support diverse tasks, from reference-to-video and identity-preserving video generation to PC-AVG.
Examples of the structured annotations are provided in Sup.~\ref{sup:Adspark-300K}.

\subsection{Data Analysis}
\input{tex/statistics}
We summarize the temporal, audio, and visual characteristics of AdSpark-300K in Fig.~\ref{fig:statistics}.
AdSpark-300K spans diverse advertisement durations while exhibiting rich multi-shot structures, with over 85\% of samples containing multiple shots, primarily 2-shot (47.3\%) and 3-shot (23.6\%) compositions (Fig.~\ref{fig:statistics}(a,c)).
To characterize narration patterns, we define Audio Density ($A_{\mathrm{den}}$) as the number of spoken words per second,
i.e., $A_{\mathrm{den}} = N_{\mathrm{word}} / T$, where $N_{\mathrm{word}}$ denotes the number of spoken words and $T$ is the video duration in seconds.
Its distribution reveals diverse narration paces across the dataset (Fig.~\ref{fig:statistics}(b)). Finally, the MUSIQ distribution indicates the high visual quality of AdSpark-300K (Fig.~\ref{fig:statistics}(d)).
Further analyses are provided in the Sup.~\ref{sup:Adspark-300K}.

\section{AdSpark-Bench}

AdSpark-Bench is a comprehensive and diagnostic benchmark designed to evaluate video generation models for PC-AVG. It consists of a carefully curated held-out test set and a hierarchical evaluation framework tailored to the requirements of product-centric advertisements. Specifically, AdSpark-Bench evaluates generated advertisements across six complementary dimensions: visual quality, product fidelity, instruction adherence, temporal coherence, audio alignment, and advertisement effectiveness. 
In this section, we first introduce the benchmark statistics and then present the hierarchical evaluation metrics.

\subsection{Benchmark Statistics}
AdSpark-Bench contains 220 product-conditioned test cases covering all 40 product categories and 220 distinct sub-categories. All benchmark samples are SKU-disjoint from the AdSpark-300K training set. Each case consists of a product reference image and structured advertisement annotations specifying product identity, selling points, creative plans covering scene, style, and shot design, and aligned audio scripts. 
AdSpark-Bench is carefully curated to support advertisement-specific and shot-aware evaluation. 
It contains 45 one-shot, 131 two-shot, and 44 three-shot cases, yielding 175 multi-shot cases for evaluating cross-shot product consistency, transition quality, and narrative coherence. 
All cases are manually reviewed to ensure reference quality, annotation completeness, and generation suitability.

\subsection{Hierarchical Evaluation Metrics}

AdSpark-Bench evaluates generated advertisements across six dimensions, each comprising multiple diagnostic submetrics. We next describe the evaluation methodology for these dimensions and submetrics. More detailed metric definitions and implementation procedures are provided in Sup.~\ref{sup:Adspark-Bench}.

\paragraph{Shot-aware Evaluation.}
Direct evaluation of multi-shot videos may confuse intentional transitions with temporal artifacts and obscure shot-level generation quality. We therefore use TransNetV2~\cite{soucek2024transnet} to segment each video into ordered shots and greedily match them to the planned shots by temporal IoU. 
The resulting shot correspondences are shared across shot-structure, shot-execution, and cross-shot consistency evaluations, enabling reliable assessment of both intra-shot quality and inter-shot transitions.

\paragraph{Visual Quality.}
We follow OpenS2V~\cite{yuan2026opens2v} and adopt Aesthetic Score to assess visual appeal and MUSIQ~\cite{ke2021musiq} to evaluate perceptual image quality.

\paragraph{Product Fidelity.}
Conventional subject-consistency metrics compare complete frames with reference images, making them sensitive to background variations and inadequate for fine-grained product identity evaluation. To address this limitation, we use GroundingDINO~\cite{groundingdino} and SAM2~\cite{sam2} to extract foreground regions at different granularities, including full products, key identity regions and packaging text. Based on these regions, we design four complementary metrics:
\textbf{(1) Subject Consistency} measures cosine similarity between DINOv3~\cite{simeoni2025dinov3} features of full-product crops from video frames and the reference image, evaluating overall product appearance preservation.
\textbf{(2) Key-Region Similarity} evaluates identity-defining regions that may be overlooked by holistic features, including logos, brand marks, and packaging detail. Annotated region descriptions are used to localize corresponding regions in reference and generated frames, which are then compared using DINOv3 features.
\textbf{(3) Text Fidelity} evaluates whether annotated product-related text, such as brand names and packaging text, remains recognizable in generated videos. RapidOCR is applied to localized product crops, and each target string is matched with recognized text using minimum edit distance.
\textbf{(4) Cross-shot Product Consistency} measures product identity stability across shot transitions by computing DINOv3 feature similarity between full-product crops before and after each cut, capturing abrupt identity changes overlooked by whole-video averaging.

\paragraph{Instruction Adherence.}
Existing video-text alignment metrics mainly measure global semantic correspondence, overlooking planned shot structure, camera design, and selling-point realization. 
We therefore evaluate instruction adherence using three fine-grained shot-aware dimensions, complemented by a global video--text alignment metric:
\textbf{(1) Shot Structure Alignment} assesses whether the generated shot sequence follows the planned structure. It measures shot-count accuracy by comparing detected and planned shot numbers, while boundary accuracy evaluates whether planned shot transitions are detected within a temporal tolerance.
\textbf{(2) Shot Execution Alignment} evaluates the consistency of each matched shot with its prescribed shot type and motion design. For each matched shot, GPT-5.5 takes sampled video frames and optical-flow-based motion cues as input to assess alignment between the generated camera behavior and the planned design.
\textbf{(3) Content Alignment} focuses on whether generated videos realize the advertisement plan, including scene, style, and selling-point realization. For scene and style alignment, GPT-5.5 assesses representative frames with corresponding annotations. 
For selling-point realization, we use denser shot-aware sampling to better capture short functional demonstrations or interaction actions, and evaluate them against the expected visual realizations and success criteria.
We additionally report \textbf{(4) GmeScore}, computed with gme-Qwen2-VL-7B-Instruct~\cite{gme}, to measure global alignment between the complete advertisement prompt and sampled video frames. 

\paragraph{Temporal Coherence.}
Standard temporal metrics mainly focus on generic frame-level consistency and may overlook product-specific instability or misinterpret intentional shot transitions as temporal artifacts. To better capture these factors, we evaluate temporal coherence from three perspectives: intra-shot motion quality, transition naturalness, and temporal consistency.
\textbf{(1) Intra-shot Motion Quality} evaluates motion dynamics within each shot. We include \textit{Motion Amplitude} following OpenS2V~\cite{yuan2026opens2v} and measure \textit{Motion Smoothness} using the coefficient of variation (CV) of frame-wise motion magnitudes derived from optical flow. To account for product-centric generation, we use GroundingDINO and SAM2 to track the advertised product and measure \textit{Product Motion Stability} based on mask-centroid displacement, mask-area variation, and adjacent-mask IoU.
\textbf{(2) Transition Naturalness} measures whether detected cuts form visually plausible transitions without abrupt visual artifacts. 
We first apply a pixel-level validity check to identify corrupted frames, and then use GPT-5.5 to assess transition naturalness in terms of brightness, style continuity, and artifact absence.
\textbf{(3) Temporal Consistency} evaluates local temporal stability around shot boundaries. For each detected boundary, we inspect duplicated, flickering, or structurally corrupted frames using adjacent-frame differences and SSIM~\cite{nilsson2020understanding}.

\paragraph{Audio Alignment.}
For videos containing a valid non-silent audio stream, we evaluate audio alignment from two aspects. 
\textbf{(1) Background Sound Consistency} uses LAION-CLAP~\cite{wu2023large} to measure the semantic similarity between the generated audio and the expected background sound description specified in the creative plan.
\textbf{(2) Narration Script Consistency} uses Whisper-Large-v3-Turbo~\cite{radford2023robust} to transcribe the generated narration and compares the recognized text with the reference audio script using character error rate (CER), measuring whether the intended commercial message is faithfully conveyed.

\paragraph{Advertisement Effectiveness.}
A visually coherent video may still fail as an advertisement if it lacks viewer appeal, commercial persuasiveness, or coherent storytelling. We therefore employ GPT-5.5 to serve as a potential customer to assess advertisement effectiveness across three aspects.
\textbf{(1) Advertisement Attractiveness} evaluates whether the generated advertisement can capture viewer attention and stimulate purchase interest from a customer perspective, considering visual appeal, product desirability, and overall viewing experience.
\textbf{(2) Creative Quality} evaluates the artistic and commercial quality of generated advertisements, including visual impact, commercial readiness, and product-oriented creativity.
\textbf{(3) Narrative Coherence} evaluates whether multi-shot advertisements achieve coherent storytelling rather than disconnected scenes, considering shot-to-shot consistency, selling-point progression, narrative logic, and pacing.

%% file: tex/pipeline.tex
\begin{figure*}
    \centering
    \includegraphics[width=\linewidth]{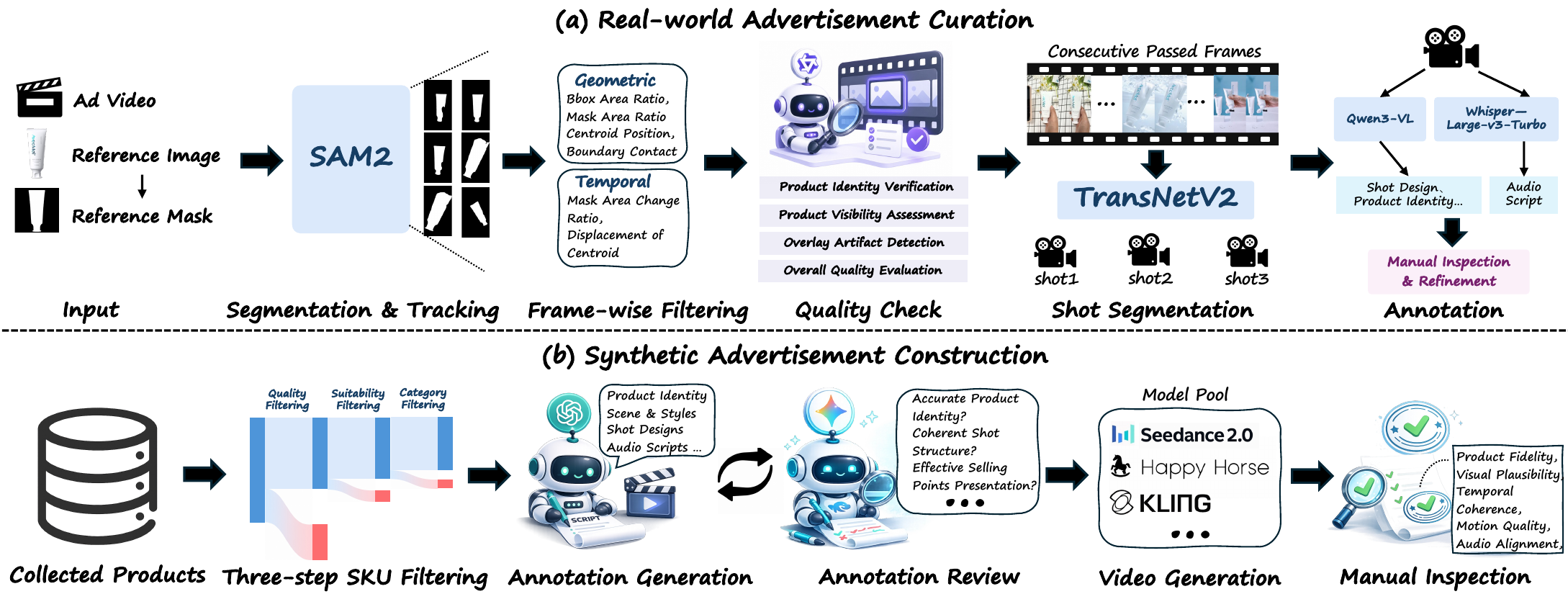}
    \caption{\textbf{Overview of the AdSpark-300K construction pipeline,} comprising real-world advertisement curation and synthetic advertisement construction, each involving tailored stages for comprehensive and unified annotation.}
    \label{fig:data_pipeline}
\end{figure*} 

%% file: tex/statistics.tex
\begin{figure*}
    \centering
    \includegraphics[width=\linewidth]{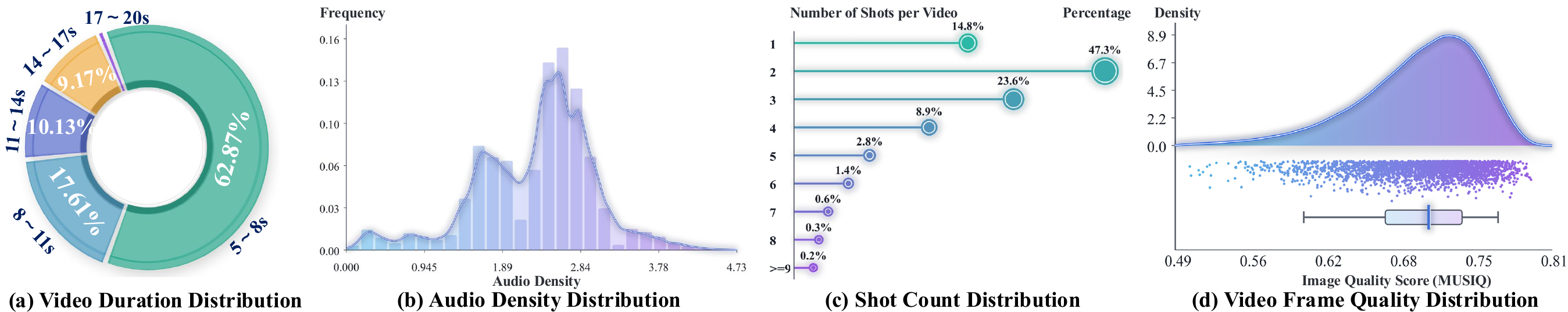}
    \caption{Statistics of AdSpark-300K, including (a) video duration distribution, (b) audio density distribution, (c) shot count distribution, and (d) video frame quality distribution via MUSIQ~\cite{ke2021musiq}, with MUSIQ scores normalized to [0, 1]. 
    Additional statistics are provided in Fig.~\ref{fig:additional_dataset_statistics}.}
    \label{fig:statistics}
\end{figure*}

%% file: Sections/4_experiments.tex
\section{Experiments}
\input{Tables/exp_2}
\subsection{Experimental Settings}

\paragraph{Baseline.}
We evaluate a set of video generation models on AdSpark-Bench, covering both closed-source and open-source approaches, including ViduQ2, ViduQ3, Seedance 2.0~\cite{seedance2026seedance}, HappyHorse-1.1, Pixverse V5, Pixverse V6, Kling 3.0 Omni~\cite{team2025kling}, VACE-14B~\cite{jiang2025vace}, SkyReels-V3-14B~\cite{li2026skyreels}, Phantom-14B~\cite{liu2025phantom}, VINO-13B~\cite{chen2026vino}, Refalign-14B~\cite{wang2026refalign}, Kaleido-14B~\cite{zhang2025kaleido}, HunyuanCustom-13B~\cite{hu2025hunyuancustom}, Bernini-14B~\cite{team2026bernini}, MV-S2V-14B~\cite{song2026mv}, MAGREF-14B~\cite{deng2025magref}, and LTX-2-14B~\cite{hacohen2026ltx}. 
\paragraph{Implementation Details.}
For fair comparison, closed-source models are evaluated through their official interfaces, while open-source models use officially released weights with recommended inference settings. All models are evaluated on the AdSpark-Bench test set using the proposed metrics. 
We further fine-tune LTX-2 on AdSpark-300K with LoRA (rank 64), denoted as LTX-AdSpark, to validate the effectiveness of our dataset for PC-AVG. All experiments are conducted on 8 NVIDIA B200 GPUs, with additional details provided in Sup.~\ref{sup:detail_experiment}.

\input{tex/qualitative_case}
\subsection{Evaluation Results}
\paragraph{Quantitative Evaluation.}

We report quantitative comparisons on AdSpark-Bench in Tab.~\ref{table_quantitative}. Existing video generation models achieve competitive performance in visual quality, but fall short in some advertisement-specific capabilities. We summarize three main observations from the results.
(1) Closed-source models generally outperform open-source models, yet still face some notable challenges. For example, although they can generate more dynamic and visually engaging videos, preserving product fidelity during complex motions remains a prominent issue, with motion blur, shape distortion, and appearance drift degrading subject consistency, key-region similarity and text fidelity.
(2) Open-source models exhibit more pronounced limitations in instruction adherence and advertisement effectiveness, and often struggle with coherent multi-shot generation, as reflected by weaker cross-shot product consistency and temporal coherence.
(3) Fine-tuning with AdSpark-300K substantially improves PC-AVG performance. Compared with the base LTX-2 model, LTX-AdSpark improves cross-shot product consistency from 33.30 to 93.59, shot structure alignment from 46.29 to 77.53, and narrative coherence from 33.77 to 89.89, demonstrating the utility of AdSpark-300K for PC-AVG.
We further analyze the effects of real-world and synthetic data composition and training scale through controlled ablations in Tab.~\ref{tab:data_composition_ablation}.

\paragraph{Qualitative Comparison.}
We provide qualitative comparisons on AdSpark-Bench in Fig.~\ref{fig:qualitative_exp}. 
Existing video generation models can produce relatively visually plausible videos while still suffering from product-specific issues, including color shifts, product deformation, and instruction deviations.
Consistent with the quantitative results, closed-source models achieve overall better performance.
Compared with LTX-2, LTX-AdSpark generates more product-consistent videos with improved shot-level coherence and more faithful product presentation, further validating that fine-tuning on AdSpark-300K effectively adapts video generation models to PC-AVG.
More qualitative results and visualizations are presented in Sup.~\ref{sup:detail_experiment}.

%% file: Tables/exp_2.tex
\begin{table*}[t]
\centering
\small
\caption{Quantitative comparison of state-of-the-art video generation models on AdSpark-Bench, covering proprietary, open-source, and AdSpark-finetuned models. The best results are highlighted in \textbf{bold}, while the second-best results are \underline{underlined}. Submetric abbreviations correspond to the metrics introduced in the benchmark section, in the same order. Dashes (--) in the audio alignment columns indicate unavailable or invalid audio outputs and are excluded from evaluation.}
\label{table_quantitative}
\setlength{\tabcolsep}{3pt}
\scalebox{0.66}{
\begin{tabular}{l|cc|cccc|cccc|ccc|cc|ccc}
\toprule

& \multicolumn{2}{c|}{\textbf{Visual Quality}}
& \multicolumn{4}{c|}{\textbf{Product Fidelity}}
& \multicolumn{4}{c|}{\textbf{Instruction Adherence}}
& \multicolumn{3}{c|}{\textbf{Temporal Coherence}}
& \multicolumn{2}{c|}{\textbf{Audio Alignment}}
& \multicolumn{3}{c}{\textbf{Ad Effectiveness}}
\\

\cmidrule(lr){2-3}
\cmidrule(lr){4-8}
\cmidrule(lr){9-11}
\cmidrule(lr){12-14}
\cmidrule(lr){15-16}
\cmidrule(lr){17-19}

\textbf{Method}
& Aes.
& Img.
& Subj.
& Reg.
& Text
& X-shot
& Struct.
& Exec.
& Content
& Gme
& Motion
& Trans.
& Temp.
& BG Audio
& Script
& Attr.
& Creat.
& Narr.
\\
\midrule

ViduQ2
&34.56&69.63
& 70.16 & 66.49 & 25.69 & 2.85
&35.77&63.75&77.94&48.99
&57.44&2.73&2.86
&--& --
&62.40&55.80&3.02
\\

ViduQ3
&38.26&70.68
& 60.57 & 58.72 & \underline{33.60} & 84.76
&71.26&60.57&83.75&48.22
&48.93&78.19&41.62
&50.85&76.91
&62.79&55.39&81.05
\\

Seedance 2.0
&38.10&70.01
& 63.04 & 59.79 & \textbf{34.31} & \textbf{94.99}
&77.14&61.82&\textbf{85.20}&49.45
&52.03&\textbf{90.05}&84.19
&49.48&80.23
&64.11&58.57&\textbf{91.46}
\\

HappyHorse-1.1
&39.56&69.60
& 59.98 & 57.26 & 31.04 & 93.27
&\textbf{89.41}&64.22&80.43&48.69
&54.49&86.72&\textbf{95.43}
&\textbf{52.40}&\underline{83.55}
&64.94&\underline{59.00}&88.52
\\

Pixverse V5
&\underline{42.08}&\underline{74.49}
& 61.78 & 60.12 & 18.52 & 2.75
&35.48&\underline{67.21}&76.94&48.37
&\underline{57.93}&2.11&2.29
&--&--
&61.12&53.05&2.53
\\

Pixverse V6
&32.51&67.68
& 58.75 & 55.27 & 22.94 & 55.18
&50.80&62.42&83.17&47.96
&55.23&54.57&59.43
&49.77&70.13
&\textbf{65.51}&58.37&55.78
\\

Kling3.0 Omni
&38.80&68.39
& 57.62 & 53.89 & 29.14 & 90.84
&75.26&66.43&80.77&47.93
&55.61&85.86&93.90
&51.60&80.01
&61.66&55.63&87.81
\\

\midrule

VACE
&37.28&69.25
& \underline{71.22} & \underline{67.82} & 18.73 & 2.16
&35.42&45.76&72.34&49.37
&48.82&2.55&2.86
&--&--
&56.24&50.84&2.18
\\

SkyReels-V3
&35.96&\textbf{75.34}
& 62.37 & 59.63 & 19.13 & 0.55 
&35.39&47.02&70.91&48.22
&52.25&0.31&0.57
&--&--
&54.02&48.14&0.47
\\

Phantom
&38.88&69.14
& 64.26 & 62.83 & 24.20 & 6.13
&36.14&52.49&73.27&48.79
&46.37&5.19&2.86
&--&--
&57.40&52.57&5.65
\\

VINO
&34.03&70.42
& 69.55 & \textbf{67.83} & 22.65 & 0.40
&35.09&50.55&69.71&\textbf{51.43}
&\textbf{59.07}&0.49&0.57
&--&--
&53.86&48.18&0.42
\\

Refalign
&38.07&70.66
& 66.23 & 62.95 & 22.60 & 5.31 
&33.24&48.86&74.33&50.90
&54.90&4.94&5.71
&--&--
&58.45&52.32&4.48
\\

Kaleido
&41.52&72.14
& 63.82 & 61.34 & 23.15 & 58.37
&48.09&52.56&76.38&50.21
&50.06&55.60&56.86
&--&--
&62.35&56.05&53.90
\\

HunyuanCustom
&36.53&67.68
& 69.28 & 67.74 & 20.92 & 3.71
&35.55&40.50&56.34&45.90
&52.20&4.07&4.57
&--&--
&48.22&42.31&3.48
\\

Bernini
&38.88&68.70
& 60.60 & 58.89 & 29.08 & 87.47
&58.41&57.83&83.20&49.23
&56.00&83.70&91.33
&--&--
&62.18&56.41&84.15
\\

MV-S2V
&\textbf{43.44}&71.38
& \textbf{71.57} & 67.58 & 14.56 & 9.53
&37.58&51.06&74.43&50.32
&53.94&8.10&6.29
&--&--
&60.12&54.11&8.30
\\

MAGREF
&38.97&69.35
& 60.36 & 58.15 & 10.01 & 0.00
&35.03&42.53&74.31&\underline{51.21}
&52.33&0.00&0.00
&--&--
&56.30&50.74&0.00
\\

LTX-2
&35.96&66.17
& 50.39 & 47.63 & 7.59 & 33.30
&46.29&64.36&79.43&48.75
&51.29&33.66&32.67
&51.43&74.71
&63.06&56.05&33.77
\\

\midrule

LTX-AdSpark
&39.25&69.57
& 65.01 & 62.07 & 29.50 & \underline{93.59}
&\underline{77.53}&\textbf{67.48}&\underline{84.90}&49.03
&52.12&\underline{88.09}&\underline{92.00}
&\underline{51.92}&\textbf{84.41}
&\underline{65.09}&\textbf{59.29}&\underline{89.89}
\\

\bottomrule
\end{tabular}
}
\end{table*}

%% file: tex/qualitative_case.tex
\begin{figure*}[t]
    \centering
    \includegraphics[width=\linewidth]{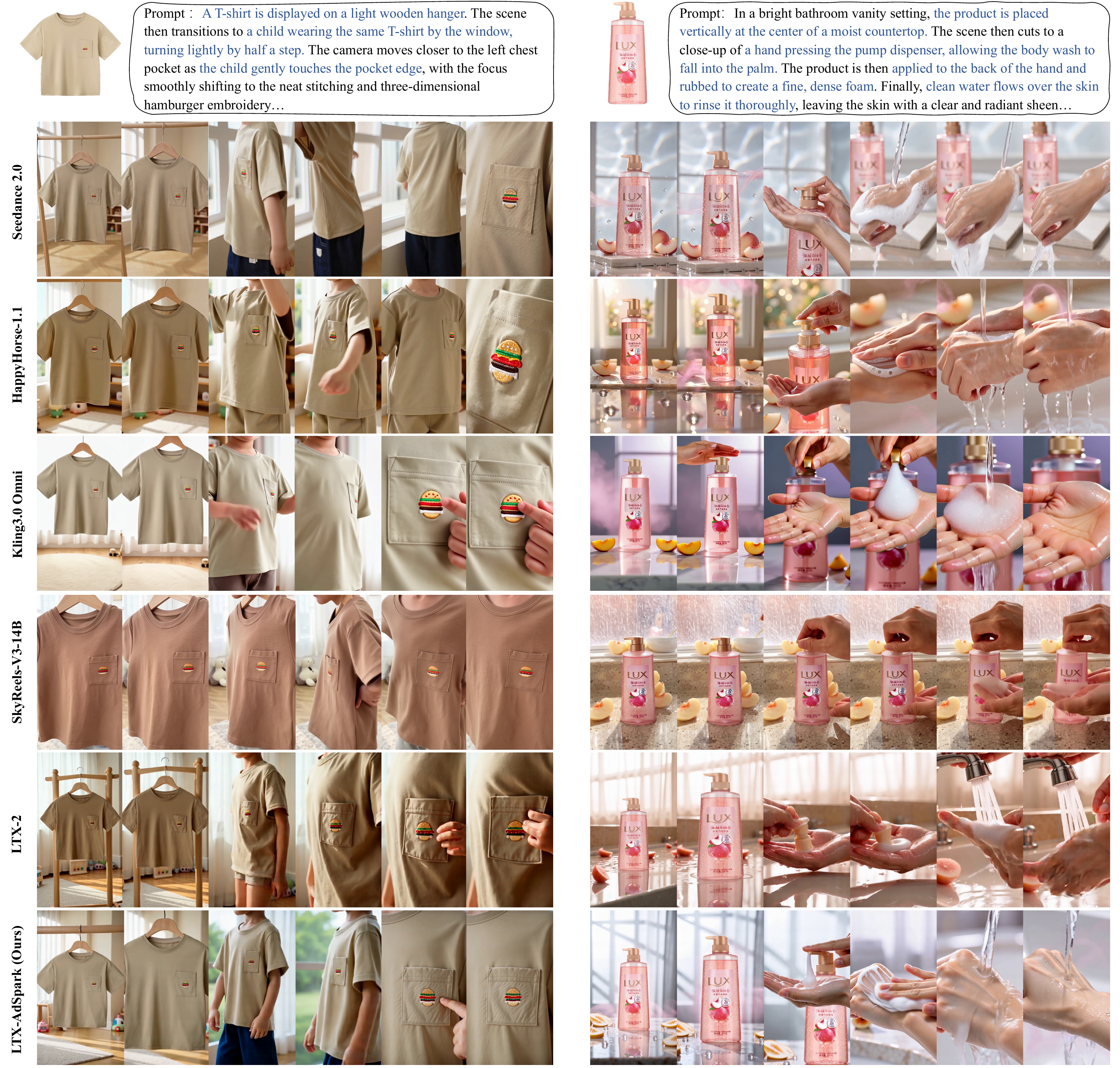}
    \caption{\textbf{Qualitative comparison on AdSpark-Bench.} We present video frames generated by different models. All methods are conditioned on the corresponding reference image and advertisement prompt. Prompts are abbreviated for readability.}
    \label{fig:qualitative_exp}
\end{figure*} 

%% file: Sections/5_conclusion.tex
\section{Conclusion}
We introduce AdSpark, a large-scale dataset and benchmark for product-centric advertisement video generation. AdSpark-300K provides 300K reference image--prompt--video triplets with structured annotations, while AdSpark-Bench evaluates six dimensions: visual quality, product fidelity, instruction adherence, temporal coherence, audio alignment, and advertisement effectiveness. Experiments show that existing models struggle with product-specific generation, while AdSpark-300K fine-tuning improves advertisement capabilities. We hope AdSpark will facilitate future research on controllable and high-quality PC-AVG.

%% file: Sections/X_supplementary.tex
\clearpage
\paragraph{Supplementary Material Overview.}
The supplementary material is organized as follows:

\begin{itemize}
  \setlength{\itemsep}{0pt}
  \setlength{\parskip}{0pt}
  \setlength{\parsep}{0pt}

  \item \textbf{Additional Details of AdSpark-300K}
  \begin{itemize}
    \setlength{\itemsep}{0pt}
    \setlength{\parskip}{0pt}
    \setlength{\parsep}{0pt}
    \item Data Sources and Collection Protocol
    \item Real-World Advertisement Curation
    \item Synthetic Advertisement Construction
    \item Unified Annotation
    \item Dataset Statistics and Distributions
    \item Representative Samples
  \end{itemize}

  \item \textbf{Additional Details of AdSpark-Bench}
  \begin{itemize}
    \setlength{\itemsep}{0pt}
    \setlength{\parskip}{0pt}
    \setlength{\parsep}{0pt}
    \item Benchmark Construction
    \item Category and Difficulty Distribution
    \item Conditional Evaluation Subsets
    \item Shot-Aware Evaluation Protocol
    \item Metric Definitions and Implementation
    \item Multimodal Judge Prompts 
    \item Score Normalization and Aggregation
    \item Metric Validation
    \item Benchmark Examples
  \end{itemize}

  \item \textbf{Additional Experimental Details}
  \begin{itemize}
    \setlength{\itemsep}{0pt}
    \setlength{\parskip}{0pt}
    \setlength{\parsep}{0pt}
    \item Details of Evaluation Models
    \item Implementation Details
    \item Qualitative Comparisons
    \item Ablation Results
  \end{itemize}

  \item \textbf{Broader Impact and Limitations}
  \begin{itemize}
    \setlength{\itemsep}{0pt}
    \setlength{\parskip}{0pt}
    \setlength{\parsep}{0pt}
    \item Limitations and Future Work
  \end{itemize}
\end{itemize}

\section{Additional Details of AdSpark-300K}
\label{sup:Adspark-300K}

\subsection{Data Sources and Collection Protocol}
\label{sup:data_source}
All data in AdSpark-300K were obtained from the product detail pages of a major e-commerce platform through authorized internal interfaces, without using external public datasets. Based on platform-wide click-through statistics on April 15, 2026, we selected approximately 600K top-ranked SKUs and retrieved all available product assets and metadata associated with them. The retrieved data may include merchant-uploaded transparent-background product images, advertisement videos with audio, SKU identifiers, product titles, brands, category labels, attributes, and selling-point descriptions. Foreground masks were derived from the alpha channels of valid transparent-background reference images. Since not every SKU is associated with all asset types, we construct the real-world and synthetic subsets using different eligibility criteria.
For the real-world subset, we retain SKUs with valid associations among an advertisement video, a transparent-background product reference image, and the corresponding product metadata. Each SKU may be associated with multiple videos, and each video may be further segmented into multiple clips. For the synthetic subset, we retain SKUs with valid reference images and product metadata, regardless of whether a paired advertisement video is available. We remove samples with invalid SKU associations, missing required assets, corrupted files, incomplete metadata, duplicate content, unsupported formats, unsuitable video durations, or sensitive product categories. The real-world and synthetic subsets may share SKUs, while AdSpark-Bench is strictly SKU-disjoint from AdSpark-300K.

\subsection{Real-World Advertisement Curation}
\label{sup:real-world-adv-cur}
\paragraph{SKU and Video Asset Filtering.}
We first identify candidate SKUs with valid product reference images, associated metadata, and at least one advertisement video. Starting from approximately 600K click-ranked SKUs, we filter the associated videos according to their duration and frame rate. Specifically, we retain videos with durations between 10 and 1,000 seconds and frame rates of 24 or 25 fps. This filtering results in 192,077 SKUs with at least one eligible advertisement video, comprising 218,837 video entries in total. Among them, 26,760 SKUs are associated with multiple eligible videos. Each retained SKU--video pair is then processed independently by the subsequent product-tracking and clip-level filtering pipeline.

\paragraph{Product Tracking and Temporal Filtering.}
For each SKU-associated advertisement video, we initialize product tracking using the foreground mask derived from the alpha channel of its transparent-background reference image. We employ SAM2.1-Hiera-Large~\cite{sam2} to propagate the product mask throughout the video and apply frame-level geometric and temporal filtering. Specifically, frames are rejected when the product bounding-box or mask area occupies less than 5\% of the frame, the mask area exceeds 80\%, or the mask approaches the frame boundary within a margin of 2\%. We further discard unstable tracking results when the mask-area ratio between adjacent frames exceeds 3.0 or the normalized centroid displacement exceeds 0.35. Consecutive valid frames are grouped into candidate clips of 5--20 seconds. Longer intervals are divided into non-overlapping clips of at most 20 seconds, while intervals and residual segments shorter than 5 seconds are discarded.

\paragraph{VLM-based Quality Filtering.}
We further evaluate each candidate clip using Qwen3-VL-235B~\cite{bai2025qwen3}, conditioned on the product reference image and up to 10 representative frames sampled from the clip. The model examines whether the advertised product remains consistent with the reference image, is sufficiently visible and complete, and is free from severe occlusion or tracking errors. It also identifies intrusive overlay artifacts, such as large floating text or graphics that obscure the product, while excluding text originally printed on the product or packaging from this criterion. A candidate clip is rejected if any sampled frame exhibits substantial product identity drift, poor visibility, severe obstruction, or prominent overlay artifacts. The complete prompt template is provided in Fig.~\ref{sup:vlm_filtering_prompt}

\paragraph{Shot Segmentation and Annotation.}
Accepted clips are segmented into ordered shots using TransNetV2~\cite{soucek2024transnet}. Representative frames from each shot, together with the reference image and associated product metadata, are then provided to Qwen3-VL-235B~\cite{bai2025qwen3} to generate structured annotations, including product identity, selling points, and creative plans covering scene, style, and shot design. The prompt template is provided in Fig.~\ref{sup:realworld_annotation_prompt}.
Whisper-Large-v3-Turbo~\cite{radford2023robust} is used to transcribe the original audio, and timestamped narration is aligned with the corresponding shots. 
Finally, a team of 20 trained annotators with experience in e-commerce product content and advertisement video inspection manually verifies both the videos and their annotations. 
The videos are distributed across the annotator pool, with each sample independently reviewed by one annotator under a unified inspection guideline.
They correct inaccurate product descriptions, shot boundaries, and transcription errors, and remove samples with unreliable annotations, product inconsistency, or insufficient overall quality. Only clips that pass this final verification are retained in the real-world subset of AdSpark-300K.

\subsection{Synthetic Advertisement Construction}
\label{sup:synthetic_construction}
We construct the synthetic subset through four stages: SKU selection, structured advertisement planning, iterative annotation review, and multi-model video generation followed by manual quality inspection.

\paragraph{Three-stage SKU Filtering.}
We begin with 600,000 click-ranked candidate SKUs collected from the e-commerce platform and construct the synthetic SKU pool through three stages of filtering.

First, \textit{quality filtering} verifies the availability and basic quality of the product assets. We retain only SKUs with valid product metadata and a transparent-background reference image. The reference image is further required to have a resolution of at least $800\times800$, a valid transparency channel, and a supported image format. We also remove corrupted files, incomplete metadata, duplicate content, and categories that are unsuitable for advertisement video generation. This stage removes 199,870 SKUs and retains 400,130 quality-qualified candidates.

Second, \textit{suitability filtering} evaluates whether the retained reference images are appropriate for product-centric advertisement video generation. We employ Qwen3-VL-235B to analyze each transparent-background product image and produce structured annotations covering generation suitability, the number of independently sellable primary subjects, the number of visible items, and potential visual contamination. We retain only images assigned a \texttt{high} generation-suitability level, containing no more than four primary subjects and four visible items, and exhibiting no promotional overlays, non-product text, watermarks, human presence, or residual scene elements. This stage removes 49,290 SKUs and retains 350,840 generation-suitable candidates.

Third, \textit{category filtering} removes SKUs from product categories unsuitable for product-centric advertisement video generation. We read the category and sub-category labels from the associated product metadata and apply a predefined category exclusion list. At the category level, we exclude non-physical products, service-oriented items, sensitive medical and healthcare products, second-hand goods, digital content, and other categories with substantial advertising-compliance or generation risks. We further remove unsuitable sub-categories that remain within otherwise valid primary categories, such as virtual products, rental and installation services, prescription-related products, examination materials, and collectible or investment-oriented items. SKUs with missing, corrupted, or unreadable category metadata are also discarded. This stage removes 2,314 SKUs and retains 348,526 category-compliant candidates.

Following the three-stage filtering, 348,526 candidate SKUs remain. We rank these candidates according to platform-wide click statistics and sequentially process them in descending order. Each candidate SKU is used to generate at most one synthetic advertisement. Samples rejected during subsequent advertisement-plan review or manual inspection are discarded, and the next ranked candidates are processed until 200K quality-controlled synthetic advertisements are collected.

\input{Tables/sup_three_stage_filter}
\paragraph{Structured Advertisement Planning.}
For each selected SKU, we employ GPT-5.5 to jointly generate structured advertisement annotations and the corresponding visual and audio instructions, conditioned on the product reference and associated metadata. Consistent with the unified representation introduced in the main paper, the annotations comprise product identity, selling-point descriptions, creative plans covering scene, style, and shot design, and aligned audio scripts. Merchant-provided selling-point descriptions serve as the primary basis for advertisement planning. GPT-5.5 translates these selling points into explicit visual realizations and observable success criteria, enabling them to be demonstrated through the generated content rather than merely described in narration.
The creative plan contains one to five temporally ordered shots, each specifying its time interval, shot type, camera motion, and corresponding selling points. The shot intervals are non-overlapping and jointly cover the complete video duration. The output further includes executable shot-by-shot visual instructions and shot-aligned audio-generation instructions. GPT-5.5 is explicitly instructed to avoid introducing unsupported functions, numerical claims, or product properties. The condensed prompt used for structured annotation and visual--audio instruction generation is provided in Fig.~\ref{sup:synthetic_planning_prompt}.

\input{tex/statistics_supplementary}
\paragraph{Iterative Annotation Review.}
To improve annotation accuracy and generation feasibility, we employ Gemini-3.1-Pro-Preview as an independent reviewer. Given the product images, metadata, and generated advertisement annotations, the reviewer examines product identity accuracy, selling-point validity, consistency between selling points and their visual realizations, shot-structure coherence, physical plausibility, and visual--audio alignment.
If the annotations are rejected, GPT-5.5 revises them according to the reviewer feedback. We perform at most five review-and-refinement rounds. The annotations are accepted immediately once they pass the reviewer; otherwise, the corresponding SKU is discarded after the fifth unsuccessful round. This process removes samples containing incorrect product descriptions, unsupported selling points, contradictory shot designs, physically implausible interactions, or mismatched visual and audio instructions. The condensed reviewer prompt used in this process is provided in Fig.~\ref{sup:synthetic_review_prompt}.

\paragraph{Multi-model Video Generation.}
The accepted structured plans are used to synthesize product-centric advertisement videos. Seedance 2.0 serves as the primary generator and accounts for the majority of the synthetic subset. We additionally include HappyHorse and Kling to increase visual diversity and reduce dependence on the generation characteristics of a single model.
All models receive the product reference image and the corresponding advertisement instructions. Videos are generated in a vertical $9{:}16$ format at 720p resolution, with durations ranging from 5 to 15 seconds. Audio generation is enabled to synthesize narration, background sound, and action-related sound effects according to the aligned audio plan. The generation instructions prohibit subtitles, promotional overlays, watermarks, interface elements, and other non-product text.

\paragraph{Manual Quality Inspection.}
Each generated advertisement is reviewed by one trained annotator from the team described in Sup.~\ref{sup:real-world-adv-cur}, following a sample-level accept-or-reject protocol.
The inspection covers five aspects: product fidelity, visual plausibility, motion and temporal quality, selling-point realization, and audio quality and alignment.

(1) For product fidelity, the generated product is required to preserve the overall shape, proportions, colors, materials, packaging layout, logo, brand marks, and other identity-defining regions of the reference image. Videos exhibiting substantial deformation, color drift, packaging replacement, logo corruption, or cross-shot identity changes are rejected.

(2) For visual plausibility, the product, scene, and human--product interactions must remain visually and physically reasonable. We exclude samples containing severe rendering artifacts, malformed hands, object penetration, unsupported floating, inconsistent geometry, implausible liquid behavior, or incorrect spatial relations.

(3) For motion and temporal quality, camera motion and object motion should remain smooth and stable within each shot, while transitions between shots should be visually natural. Videos with severe flickering, duplicated or corrupted frames, abrupt appearance changes, unstable product motion, or incoherent transitions are discarded.

(4) For selling-point realization, the generated content should visibly demonstrate the intended product function, characteristic, interaction, or commercial appeal specified by the plan. A sample is rejected when its central selling points are omitted, only expressed through narration, contradicted by the visual content, or replaced by unrelated actions.

(5) For audio quality and alignment, narration and sound effects should be intelligible, semantically consistent with the planned script, and temporally aligned with the corresponding visual content. We reject videos with missing audio, severely corrupted speech, unrelated narration, substantial audio--visual mismatch, or sound effects inconsistent with the depicted actions.

We additionally discard videos containing prominent unintended subtitles, watermarks, promotional overlays, sensitive information, or other non-product text. A video is retained only when it satisfies all five inspection aspects. The resulting synthetic subset contains 200K quality-controlled product-centric advertisement videos.

\subsection{Unified Annotation}
\label{sup:unified_annotation}

Despite being constructed through different pipelines, the real-world and synthetic subsets share the same structured annotation format. As shown in Fig.~\ref{sup:unified_annotation_schema}, each sample contains five components: product identity, selling points, a creative plan, an aligned audio script, and generation prompts. This unified representation allows the two subsets to be jointly used for product-centric advertisement video generation while preserving the provenance and temporal structure of each sample.

\paragraph{Product Identity.}
The product-identity annotation describes both semantic identity and fine-grained visual appearance. It includes the product category, brand, and product name, together with visual attributes such as dominant colors, materials, shape, key identity regions, recognizable packaging text, and essential details that should remain unchanged throughout the video. The \texttt{key\_regions} field records identity-defining areas such as logos, brand marks, characteristic patterns, and packaging components, while \texttt{ocr\_targets} specifies product-related text that should remain recognizable. The \texttt{must\_keep} field summarizes the visual elements most critical to preserving product identity.

\paragraph{Selling Points.}
Each selling-point entry contains the commercial feature to be communicated, its intended visual realization, and an observable success criterion. Rather than representing selling points only as textual claims, the \texttt{visualization} field describes how they should appear through product interactions, functional demonstrations, material details, scene design, or visible effects. The \texttt{visual\_success\_criteria} field further specifies concrete visual evidence that can be assessed from the video, enabling both generation supervision and selling-point-aware evaluation.

\paragraph{Creative Plan.}
The creative plan organizes the advertisement at the scene, style, and shot levels. The scene annotation describes the environment and contextual presentation of the product, while the style annotation specifies the intended visual tone, including lighting, color palette, contrast, texture, and commercial aesthetics. Both fields include observable success criteria to reduce ambiguity.
The shot sequence provides a temporally ordered description of the advertisement narrative. Each shot is associated with a unique identifier, time interval, shot type, camera-motion type, visual description, intent, and visual success criterion. The \texttt{selling\_point\_indices} field explicitly links each shot to the selling points it presents. This linkage makes it possible to determine not only whether a selling point is included in the advertisement plan, but also when and how it is visually communicated. Shot intervals are ordered, non-overlapping, and jointly cover the complete video duration.

\paragraph{Aligned Audio Script.}
The audio script is organized according to the shot sequence. Each entry contains a shot identifier, temporal interval, background sound, and narration. This shot-level alignment associates spoken commercial messages and relevant sound effects with the corresponding visual content, supporting audio-conditioned generation and fine-grained audio--visual evaluation.
For real-world advertisements, narration is derived from timestamped ASR transcripts and aligned with detected shots. For synthetic advertisements, the narration and sound instructions are jointly planned with the visual content.

\paragraph{Generation Prompts.}
In addition to the structured fields, each sample provides executable video and audio generation prompts. The video-generation prompt converts the creative plan into a coherent shot-by-shot description of the scene, composition, camera behavior, product interaction, and selling-point presentation. The audio-generation prompt specifies shot-aligned narration and sound effects. These prompts serve as model-ready textual conditions, while the structured annotations retain fine-grained supervision and support diagnostic evaluation.

\paragraph{Annotation Alignment Across Subsets.}
The two subsets differ in how their annotations are obtained but not in their final representation. For real-world samples, product identity, selling points, scene, style, and shot designs are inferred from the reference image, sampled video frames, product metadata, detected shot boundaries, and timestamped transcript. The annotations therefore describe content already present in the source advertisement. For synthetic samples, the same fields are planned from the product reference and metadata before video generation, and the corresponding visual and audio prompts are generated jointly with the structured annotations.
All annotations are normalized to the schema in Fig.~\ref{sup:unified_annotation_schema}. We validate required fields, shot ordering, temporal coverage, selling-point references, and cross-field consistency before including a sample in AdSpark-300K. This shared representation enables unified training across real-world and synthetic data and supports tasks ranging from reference-conditioned video generation and product-identity preservation to selling-point visualization, multi-shot advertisement planning, and audio--visual generation.

\subsection{Dataset Statistics and Distributions}
\label{sup:dataset_statistics}

We further analyze the video content of AdSpark-300K in Fig.~\ref{fig:additional_dataset_statistics}. At the shot level, tracking shots and dolly-in motions dominate the camera-motion distribution, accounting for 38.1\% and 35.2\%, respectively, followed by rack focus at 13.6\%. This indicates that the dataset primarily favors smooth subject following and gradual visual emphasis, which are well suited to product-centric presentation. In terms of shot types, demonstration shots are the most frequent, followed by hero shots, macro-detail shots, and full-product shots, reflecting the importance of functional visualization and fine-grained product presentation in advertisement videos. Most products are associated with two or three selling points, while cases with more than three are relatively rare, suggesting that the advertisements generally focus on a compact set of core commercial messages. The prompt-length distribution peaks at around 500 characters while covering a wide range, reflecting diverse levels of scene detail, shot complexity, and selling-point presentation across the dataset.

\input{tex/benchmark_product}
\subsection{Representative Samples}
\input{tex/dataset_presentation}
\input{tex/annotation}

We present representative samples from AdSpark-300K across a wide range of product categories in Fig.~\ref{fig:dataset_presentation}, where each reference image is paired with advertisement video frames sampled in temporal order. These examples demonstrate the diversity of product appearances, advertising scenes, camera motions, and product-oriented interactions covered by the dataset, while also showing how product identity and key selling points are preserved throughout the generated videos.
We further provide examples of advertisement videos together with their structured prompts in Fig.~\ref{fig:example_annotatio_video}. The prompts describe the overall scene and visual style, shot-level composition and camera motion, temporal segmentation, and synchronized audio cues, illustrating how commercial intentions are translated into executable multi-shot plans. A complete annotation for the diving-watch sample is additionally shown in Fig.~\ref{sup:annotation_example_adspark_300k}, providing a concrete example of the unified annotation schema used throughout AdSpark-300K.

\section{Additional Details of AdSpark-Bench}
\label{sup:Adspark-Bench}

\subsection{Benchmark Construction}
AdSpark-Bench is constructed from held-out product SKUs to provide a reliable evaluation suite for product-centric advertisement video generation. We randomly select 350 SKUs from the eligible product pool while ensuring that all benchmark samples are strictly SKU-disjoint from the AdSpark-300K training set. Each selected SKU contains a high-quality reference image, complete advertisement annotations, and structured generation conditions, including product identity, selling points, creative plans, and audio scripts.
To ensure evaluation reliability, we apply several filtering criteria during benchmark construction. We retain products with a single dominant object, high visual quality, clear product identity, and complete annotation schemas. Samples with low-resolution reference images, ambiguous product categories, missing metadata, or unsuitable generation conditions are removed through manual inspection. Furthermore, we restrict the benchmark to one sample per fine-grained product category (SKU category level) to reduce redundancy and improve evaluation diversity.
The final benchmark contains 220 product-conditioned cases covering 40 product categories and 220 distinct sub-categories. Following the planned advertisement structure, the benchmark includes 45 one-shot, 131 two-shot, and 44 three-shot cases, resulting in 175 multi-shot cases for evaluating cross-shot consistency, transition quality, and narrative coherence.

\subsection{Category and Difficulty Distribution}
AdSpark-Bench is designed to cover diverse product categories and generation difficulties. As shown in Fig.~\ref{fig:benchmark_product}, the benchmark spans all 40 product categories in AdSpark-300K, including electronics, beauty products, clothing, furniture, kitchenware, and beverages. Category selection follows the natural distribution of available products while maintaining coverage for long-tail categories through category-level balancing.
Beyond category diversity, we explicitly consider generation difficulty factors that are critical for product-centric advertisement generation. The benchmark contains diverse visual challenges, including fine-grained product details, packaging text preservation, logo recognition, multi-shot storytelling, and product-motion interaction. We further maintain a balanced shot-complexity distribution, where one-shot cases evaluate basic product preservation, while two-shot and three-shot cases introduce additional challenges in shot transition, temporal consistency, and narrative progression.
These design choices enable AdSpark-Bench to evaluate not only general video generation quality, but also the capability of models to preserve product identity and realize advertisement-oriented creative intentions under different levels of difficulty.

\subsection{Conditional Evaluation Subsets}
Some advertisement-specific metrics require specific visual conditions to provide reliable measurements. Therefore, AdSpark-Bench adopts a conditional evaluation protocol, where metrics are computed on the complete benchmark whenever applicable and on dedicated subsets when additional prerequisites are required. 
Specifically, we define three conditional subsets:
\textbf{Text Fidelity Subset.}
Text Fidelity requires recognizable product-related text, such as brand names or packaging descriptions. We construct a text evaluation subset containing 72 samples with sufficient and readable OCR targets. These samples satisfy the criteria of having at least four annotated OCR targets and clear packaging text visibility.
\textbf{Logo and Key-Region Subset.}
Logo and Key-Region Similarity focuses on identity-defining visual details, including logos, brand marks, and distinctive product regions. We construct this subset with 172 samples containing visible logos or clear key regions. Specifically, samples are selected based on product identity clarity, visible brand logos, and sufficient key-region annotations.
\textbf{Product Demonstration Subset.}
To evaluate challenging advertisement scenarios involving product usage and human-product interaction, we additionally construct a product demonstration subset containing 20 samples. These cases include functional demonstrations, usage scenarios, or interaction-based shots, where models are required to visualize product functions rather than only preserve product appearance.
For multi-shot related metrics, including Cross-shot Product Consistency, Transition Naturalness, and Narrative Coherence, we directly use the multi-shot subset consisting of 175 samples (131 two-shot and 44 three-shot cases). This conditional evaluation design avoids unreliable measurements caused by missing evaluation targets while enabling fine-grained analysis of model capabilities under different advertisement scenarios.

\subsection{Shot-Aware Evaluation Protocol}
Multi-shot advertisement videos contain intentional shot transitions, camera movements, and scene changes. Directly evaluating the entire video sequence may incorrectly treat designed transitions as temporal artifacts and obscure shot-level generation quality. Therefore, AdSpark-Bench adopts a shot-aware evaluation protocol to analyze shot structure, execution, temporal coherence, and cross-shot consistency.

Given a generated video $V$, we first apply TransNetV2~\cite{soucek2024transnet} to detect shot boundaries and divide the video into an ordered sequence of generated shots:
\begin{equation}
\mathcal{S}^{g}=\{s^{g}_{1},s^{g}_{2},...,s^{g}_{N_g}\},
\end{equation}
where $N_g$ denotes the number of detected shots. Each generated shot $s_i^g$ is represented by its temporal interval $[t_i^s,t_i^e]$.

The detected shots are then matched with the planned shots from the advertisement creative plan:
\begin{equation}
\mathcal{S}^{p}=\{s^{p}_{1},s^{p}_{2},...,s^{p}_{N_p}\},
\end{equation}
where each planned shot contains structured information including shot type, motion design, and visual objectives.

To establish shot correspondence, we compute temporal Intersection-over-Union (tIoU) between generated and planned shots:
\begin{equation}
\mathrm{tIoU}(s_i^g,s_j^p)
=
\frac{|s_i^g \cap s_j^p|}
{|s_i^g \cup s_j^p|},
\end{equation}
where $|\cdot|$ denotes the temporal duration of an interval. We greedily match each generated shot with the unmatched planned shot having the highest tIoU:
\begin{equation}
j^*=\arg\max_j \mathrm{tIoU}(s_i^g,s_j^p).
\end{equation}

The obtained shot correspondences are shared across multiple evaluation dimensions. Specifically, shot structure alignment evaluates whether the generated shot sequence follows the planned organization, while shot execution alignment measures whether each matched shot realizes the intended camera behavior and motion design. Temporal metrics further evaluate motion quality within individual shots and transition quality across consecutive shots.

Importantly, shot detection is performed solely based on the generated video without using ground-truth shot boundaries. This schema-blind protocol prevents annotation leakage and ensures that shot-related metrics reflect the actual generation capability of each model. For single-shot advertisements, transition-related metrics are skipped, while all other applicable metrics remain unchanged.

\subsection{Metric Definitions and Implementation}

\subsubsection{Visual Quality}

Visual quality evaluates the perceptual quality of generated advertisements without considering product identity or instruction compliance. We adopt two complementary metrics, including Aesthetic Score and MUSIQ~\cite{ke2021musiq}, to assess visual attractiveness and image quality, respectively.

Given a generated video $V$, we uniformly sample $N$ frames:
\begin{equation}
\mathcal{F}=\{f_1,f_2,\dots,f_N\},
\end{equation}
where $N=16$ in our experiments. For each sampled frame, we independently compute the aesthetic score and imaging quality score. The final metric value is obtained by averaging frame-level scores:
\begin{equation}
M(V)=\frac{1}{N}\sum_{i=1}^{N}M(f_i),
\end{equation}
where $M(\cdot)$ denotes either Aesthetic Score or MUSIQ.

Aesthetic Score is normalized from its original range to $[0,1]$, while MUSIQ directly produces a normalized image quality score. We report these two metrics separately to characterize different aspects of perceptual video quality.

\subsubsection{Product Fidelity}

Product fidelity measures whether generated advertisements faithfully preserve the identity and appearance of the referenced product. Different from conventional subject consistency metrics that rely on holistic image similarity, AdSpark-Bench evaluates product fidelity at multiple granularities, including overall product appearance, identity-defining regions, packaging text, and cross-shot stability.

Given a reference image $I_r$, generated video $V$, and product annotations $\mathcal{A}$, we first extract product-related regions using GroundingDINO~\cite{groundingdino} and refine the corresponding masks using SAM2~\cite{sam2}. The extracted regions are then encoded with DINOv3~\cite{simeoni2025dinov3} for feature-level comparison. The complete evaluation procedure is summarized in Algorithm~\ref{alg:product_fidelity}.

\input{Algorithm/alg_product_fidelity}

\paragraph{Subject Consistency.}
Subject Consistency evaluates whether the generated product preserves its overall appearance. Let $z_r$ denote the DINOv3 feature of the reference product region, and $z_i$ denote the feature of the generated product crop from the $i$-th sampled frame. The similarity is computed as:
\begin{equation}
S_{\mathrm{subj}}
=
\frac{1}{N}
\sum_{i=1}^{N}
\cos(z_r,z_i).
\end{equation}

\paragraph{Key-Region Similarity.}
Key-Region Similarity focuses on fine-grained identity-defining regions, such as logos, brand marks, and distinctive packaging details. Given a set of annotated key regions $\mathcal{R}$, the similarity is computed as:
\begin{equation}
S_{\mathrm{region}}
=
\frac{1}{|\mathcal{R}|}
\sum_{j\in\mathcal{R}}
\cos(z_j^r,z_j^g),
\end{equation}
where $z_j^r$ and $z_j^g$ denote the DINOv3 features of the $j$-th region in the reference and generated videos, respectively.

\paragraph{Text Fidelity.}
Text Fidelity measures whether product-related text, such as brand names and packaging information, remains recognizable after generation. Given the ground-truth text $y$ and OCR-recognized text $\hat{y}$, we compute:
\begin{equation}
S_{\mathrm{text}}
=
1-
\frac{d_{\mathrm{edit}}(y,\hat{y})}
{|y|},
\end{equation}
where $d_{\mathrm{edit}}(\cdot,\cdot)$ denotes the Levenshtein edit distance. For multiple text regions, the final score is averaged over all valid text targets.

\paragraph{Cross-shot Product Consistency.}
For multi-shot advertisements, we additionally evaluate product stability across shot transitions. 
Given adjacent shots $s_i$ and $s_{i+1}$, we compute the feature similarity between product regions around each transition:
\begin{equation}
S_{\mathrm{xshot}}
=
\frac{1}{K}
\sum_{i=1}^{K}
\cos(z_i,z_{i+1}),
\end{equation}
where $K$ denotes the number of detected shot transitions. This metric captures abrupt product identity changes that may be overlooked by frame-level averaging.

\subsubsection{Instruction Adherence}

Instruction adherence evaluates whether the generated advertisement follows the provided creative plan, including shot structure, shot-level execution, scene and style requirements, and selling-point realization. Unlike generic video-text alignment metrics, AdSpark-Bench performs shot-aware evaluation based on the detected shot sequence and the structured advertisement annotations.

Given a generated video $V$ and its corresponding creative plan $\mathcal{C}$, we first obtain detected shots using the shot-aware protocol described above and match them with the planned shots. The overall evaluation procedure is summarized in Algorithm~\ref{alg:instruction_adherence}. All instruction adherence metrics are reported independently without aggregating them into a weighted score.

\input{Algorithm/alg_instruction_adherence}

\paragraph{Shot Structure Alignment.}
Shot Structure Alignment measures whether the generated video follows the planned number and temporal organization of shots. Let $N_g$ and $N_p$ denote the numbers of detected and planned shots, respectively. We first compute shot-count accuracy:
\begin{equation}
S_{\mathrm{count}}
=
\max\left(0, 1-\frac{|N_g-N_p|}{N_p}\right).
\end{equation}
We further evaluate whether planned shot boundaries are correctly realized. Let $\mathcal{B}^{p}$ and $\mathcal{B}^{g}$ denote the planned and detected boundary sets. A planned boundary is considered matched if a detected boundary lies within a temporal tolerance $\delta=0.5$s:
\begin{equation}
S_{\mathrm{bound}}
=
\frac{1}{|\mathcal{B}^{p}|}
\sum_{\tau \in \mathcal{B}^{p}}
\mathbf{1}
\left[
\min_{b\in\mathcal{B}^{g}}|\tau-b|\leq \delta
\right].
\end{equation}
The final Shot Structure Alignment score is computed from shot-count and boundary accuracy:
\begin{equation}
S_{\mathrm{struct}}
=
0.4 S_{\mathrm{count}} + 0.6 S_{\mathrm{bound}}.
\end{equation}
For single-shot cases without internal boundaries, boundary accuracy is set to 1.

\paragraph{Shot Execution Alignment.}
Shot Execution Alignment evaluates whether each generated shot realizes the planned shot type and motion design. For each matched shot pair $(s_i^g,s_j^p)\in\mathcal{M}$, we sample frames from the generated shot and construct contact sheets. We additionally compute optical-flow cues to provide coarse motion evidence. GPT-5.5 then compares the visual evidence with the corresponding planned shot annotations, including shot type, camera behavior, and motion type. The score is computed as:
\begin{equation}
S_{\mathrm{exec}}
=
\frac{1}{|\mathcal{M}|}
\sum_{(i,j)\in\mathcal{M}}
q_{ij}^{\mathrm{exec}},
\end{equation}
where $q_{ij}^{\mathrm{exec}}\in[0,1]$ denotes the normalized GPT-5.5 score for the matched shot. Extra generated shots without matched planned shots receive zero execution scores.

\paragraph{Content Alignment.}
Content Alignment measures whether the generated advertisement realizes the intended scene, style, and selling points. We use two frame-sampling strategies. For scene and style alignment, we sample representative frames from each detected shot to capture the global visual context. For selling-point realization, we use denser shot-aware sampling to better capture short functional demonstrations or interaction actions. GPT-5.5 evaluates each component according to the structured creative plan and the expected visual success criteria. The score is computed as:
\begin{equation}
S_{\mathrm{content}}
=
0.3 S_{\mathrm{scene}}
+
0.3 S_{\mathrm{style}}
+
0.4 S_{\mathrm{sell}},
\end{equation}
where $S_{\mathrm{scene}}$, $S_{\mathrm{style}}$, and $S_{\mathrm{sell}}$ denote normalized scene, style, and selling-point alignment scores, respectively.

\paragraph{GmeScore.}
We additionally report GmeScore to measure global semantic alignment between the full advertisement prompt and the generated video. Given the prompt embedding $\psi_t(p)$ and frame embedding $\psi_v(f_i)$ from gme-Qwen2-VL-7B-Instruct, GmeScore is computed as:
\begin{equation}
S_{\mathrm{gme}}
=
\frac{1}{N}
\sum_{i=1}^{N}
\psi_t(p)^\top \psi_v(f_i).
\end{equation}
This metric complements the fine-grained shot-aware evaluation by providing a global video-text relevance score.

\subsubsection{Temporal Coherence}

Temporal coherence evaluates whether generated advertisements exhibit smooth motion, stable product dynamics, natural shot transitions, and limited temporal artifacts. Since advertisement videos often contain intentional shot cuts, all temporal metrics are computed in a shot-aware manner to avoid mistaking designed transitions for temporal inconsistency.

Given the detected shot sequence, we evaluate temporal coherence from three perspectives: intra-shot motion quality, transition naturalness, and local temporal consistency around shot boundaries. The evaluation protocol is summarized in Algorithm~\ref{alg:temporal_coherence}. All temporal coherence metrics are reported independently without aggregating them into a weighted score.

\input{Algorithm/alg_temporal_coherence}

\paragraph{Intra-shot Motion Quality.}
Intra-shot Motion Quality measures whether motion within each shot is sufficiently dynamic, smooth, and product-stable. For adjacent frames, we compute dense optical flow $u_t(x)$ and obtain the average motion magnitude:
\begin{equation}
m_t
=
\frac{1}{|\Omega|}
\sum_{x\in\Omega}
\|u_t(x)\|_2,
\end{equation}
where $\Omega$ denotes the image domain. Motion Amplitude is computed as the average frame-wise motion magnitude:
\begin{equation}
A
=
\frac{1}{T-1}
\sum_{t=1}^{T-1}m_t.
\end{equation}
Following benchmark-level normalization, raw amplitudes are clipped by the 5th and 95th percentiles and linearly mapped to $[0,1]$.

Motion Smoothness is measured using the coefficient of variation of frame-wise motion magnitudes:
\begin{equation}
S_{\mathrm{smooth}}
=
\max\left(0, 1-\frac{\sigma(m)}{\mu(m)+\epsilon}\right),
\end{equation}
where $\mu(m)$ and $\sigma(m)$ denote the mean and standard deviation of the motion magnitude sequence. Static shots with near-zero mean motion are assigned a smoothness score of 1.

To account for product-centric generation, we further track the advertised product within each shot and evaluate product motion stability. Given product masks across adjacent frames, we compute centroid stability, area stability, and mask-IoU stability. The final product stability score is:
\begin{equation}
S_{\mathrm{stab}}
=
0.4S_{\mathrm{cent}}
+
0.3S_{\mathrm{area}}
+
0.3S_{\mathrm{iou}}.
\end{equation}
The intra-shot motion score is then computed as:
\begin{equation}
S_{\mathrm{motion}}
=
0.25A_{\mathrm{norm}}
+
0.40S_{\mathrm{smooth}}
+
0.35S_{\mathrm{stab}}.
\end{equation}

\paragraph{Transition Naturalness.}
Transition Naturalness measures whether shot cuts form visually plausible transitions without abrupt artifacts. For each detected boundary, we examine frames before and after the cut. We first apply a pixel-level validity check to identify corrupted frames, such as black frames, over-exposed frames, or near-constant frames. If the boundary passes this check, GPT-5.5 further evaluates whether the transition is visually natural in terms of brightness, style continuity, and artifact absence. The score is computed as:
\begin{equation}
S_{\mathrm{trans}}
=
\frac{1}{K}
\sum_{k=1}^{K}
q_{k}^{\mathrm{trans}},
\end{equation}
where $q_{k}^{\mathrm{trans}}\in[0,1]$ is the naturalness score of the $k$-th transition, and $K$ is the number of detected shot boundaries. Single-shot videos have no transition score and are excluded from this metric.

\paragraph{Temporal Consistency.}
Temporal Consistency focuses on local temporal artifacts around shot boundaries, including duplicated frames, flickering, and structurally corrupted frames. For each detected boundary, we inspect a local temporal window and compute adjacent-frame intensity differences and SSIM values. A normal shot transition should contain at most one major visual change within the window, while repeated large changes indicate flickering or unstable transitions.

For each boundary, we assign a binary consistency score:
\begin{equation}
q_k^{\mathrm{temp}}
=
\mathbf{1}
\left[
n_{\mathrm{diff}}^{k}\leq 1
\land
n_{\mathrm{ssim}}^{k}\leq 1
\land
\neg c_k
\right],
\end{equation}
where $n_{\mathrm{diff}}^{k}$ is the number of excessive intensity jumps, $n_{\mathrm{ssim}}^{k}$ is the number of structural drops, and $c_k$ indicates whether corrupted or repeated frames are detected. The final Temporal Consistency score is:
\begin{equation}
S_{\mathrm{temp}}
=
\frac{1}{K}
\sum_{k=1}^{K}
q_k^{\mathrm{temp}}.
\end{equation}
This metric is only computed for videos with detected shot boundaries.

\subsubsection{Audio Alignment}

Audio alignment evaluates whether the generated audio is consistent with the intended advertisement design. Since not all video generation models support audio generation, this dimension is computed only for videos with valid audio outputs. We evaluate audio alignment from two complementary aspects: background sound consistency and narration script consistency.

Background Sound Consistency measures whether the generated background audio matches the expected audio description in the creative plan. Narration Script Consistency evaluates whether the spoken narration follows the provided audio script. The overall evaluation protocol is summarized in Algorithm~\ref{alg:audio_alignment}. All audio alignment metrics are reported independently.

\input{Algorithm/alg_audio_alignment}

\paragraph{Background Sound Consistency.}
We use LAION-CLAP~\cite{wu2023large} to measure the semantic consistency between the generated audio and the expected background sound description. Let \(\eta_a(\cdot)\) and \(\eta_t(\cdot)\) denote the audio and text encoders, respectively. The score is computed as:
\begin{equation}
S_{\mathrm{bg}}
=
\cos(\eta_a(a),\eta_t(d_a)),
\end{equation}
where \(a\) is the generated audio and \(d_a\) is the audio description derived from the benchmark annotation. The score is normalized to \([0,1]\) for reporting.

\paragraph{Narration Script Consistency.}
For advertisements containing narration, we transcribe the generated audio using Whisper-Large-v3-Turbo~\cite{radford2023robust} and compare the transcription with the reference audio script. Given the reference script \(y\) and the recognized script \(\hat{y}\), we compute:
\begin{equation}
S_{\mathrm{script}}
=
1-
\frac{d_{\mathrm{edit}}(y,\hat{y})}{|y|},
\end{equation}
where \(d_{\mathrm{edit}}(\cdot,\cdot)\) denotes the Levenshtein edit distance. This metric reflects whether the generated narration preserves the intended commercial message.

\subsubsection{Advertisement Effectiveness}

Advertisement effectiveness evaluates whether a generated video functions well as a product advertisement, beyond low-level visual quality and instruction following. A video may preserve the product and follow the prompt, but still fail to attract customers, present the selling point persuasively, or form a coherent advertisement narrative. Therefore, AdSpark-Bench evaluates advertisement effectiveness from a customer-oriented perspective.

We employ GPT-5.5 as a potential customer evaluator. Given the reference product, creative plan, selling points, and sampled frames from the generated video, GPT-5.5 assigns scores to three dimensions: Advertisement Attractiveness, Creative Quality, and Narrative Coherence. The evaluation protocol is summarized in Algorithm~\ref{alg:ad_effectiveness}. These three metrics are reported independently without aggregating them into an overall score.

\input{Algorithm/alg_ad_effectiveness}

\paragraph{Advertisement Attractiveness.}
Advertisement Attractiveness measures whether the generated video can capture viewer attention and stimulate purchase interest. GPT-5.5 evaluates this dimension from the perspective of a potential customer, considering visual appeal, product desirability, and the overall viewing experience. The raw score \(q_{\mathrm{attr}}\) is normalized for reporting:
\begin{equation}
S_{\mathrm{attr}}=\mathrm{Norm}(q_{\mathrm{attr}}).
\end{equation}

\paragraph{Creative Quality.}
Creative Quality evaluates whether the generated video exhibits a polished and commercially usable advertisement design. The evaluator considers visual impact, commercial readiness, product-oriented creativity, and whether the creative expression supports the intended selling points. The score is computed as:
\begin{equation}
S_{\mathrm{creat}}=\mathrm{Norm}(q_{\mathrm{creat}}).
\end{equation}

\paragraph{Narrative Coherence.}
Narrative Coherence evaluates whether multi-shot advertisements form a coherent storytelling progression rather than disconnected scenes. GPT-5.5 considers shot-to-shot consistency, selling-point progression, narrative logic, and pacing. The score is computed as:
\begin{equation}
S_{\mathrm{narr}}=\mathrm{Norm}(q_{\mathrm{narr}}).
\end{equation}

For single-shot advertisements, Narrative Coherence focuses on whether the generated content forms a complete and self-contained advertisement presentation. For multi-shot advertisements, it further considers whether the shots are logically connected and whether the selling points are progressively communicated.

\input{tex/case1}
\input{tex/case2}

\subsection{Multimodal Judge Prompts}
Several metrics in AdSpark-Bench require high-level multimodal reasoning that cannot be reliably captured by low-level visual features or automatic detectors alone. We therefore use GPT-5.5 as a multimodal judge for selected advertisement-specific metrics, including Shot Execution Alignment, Content Alignment, Transition Naturalness, and Advertisement Effectiveness. 
All multimodal judge prompts follow a unified evaluation format. The judge receives temporally ordered frames or shot-level contact sheets together with the corresponding structured annotations, such as planned shot type, motion type, scene, style, selling points, and shot descriptions. For Likert-based prompts, GPT-5.5 outputs integer scores from 1 to 5, where 1 indicates poor or missing realization, 3 indicates partial realization, and 5 indicates strong realization. These scores are linearly mapped to $[0,1]$ for reporting. For transition naturalness and narrative coherence, the judge directly outputs a scalar score in $[0,1]$. All prompts require strict JSON or numeric-only outputs to ensure reliable parsing and reproducibility. Representative prompt templates are provided in Fig.~\ref{sup:shot_execution_prompt}--Fig.~\ref{sup:narrative_prompt}.

Although GPT-5.5 is used in both synthetic advertisement planning and selected benchmark metrics, this does not constitute direct self-evaluation, as the two stages serve fundamentally different roles and operate on different inputs. During data construction, GPT-5.5 performs language-based creative planning from product references, metadata, and selling points to produce structured advertisement instructions. During evaluation, it instead performs multimodal reasoning over independently generated visual evidence, including sampled frames, shot-level contact sheets, and motion cues, to assess how well a generated video realizes a fixed product reference and advertisement plan. Therefore, the evaluation is grounded in the generated visual content rather than textual similarity to the planner output. For each benchmark case, all evaluated video generation models are conditioned on exactly the same reference image and advertisement prompt, such that the conditioning information is held constant across methods. We further validate the GPT-assisted metrics independently against human judgments in Sup.~\ref{sup:metric_val}.

\input{tex/case3}
\subsection{Score Normalization and Aggregation}

AdSpark-Bench includes heterogeneous metrics from pretrained perceptual models, feature similarity, OCR, audio-text alignment, and multimodal judge scores. To make the results comparable, all reported metrics are normalized to a common range before aggregation. Unless otherwise specified, each metric is first mapped to $[0,1]$ and then reported in percentage form.

For metrics that naturally produce bounded similarity scores, such as cosine similarity, CLAP similarity, GmeScore, and normalized edit-distance similarity, we directly use their normalized values. For raw perceptual scores with dataset-dependent ranges, we apply linear clipping and rescaling:
\begin{equation}
\mathrm{Norm}(x)
=
\mathrm{clip}\left(
\frac{x-l}{u-l}, 0, 1
\right),
\end{equation}
where $l$ and $u$ denote the lower and upper normalization bounds. In our implementation, Aesthetic Score is normalized with $l=4.0$ and $u=7.0$, while MUSIQ is divided by 100 to obtain an image-quality score in $[0,1]$.

For GPT-5.5 judge scores based on a 1--5 Likert scale, we linearly map the raw score $s$ to $[0,1]$:
\begin{equation}
\mathrm{Norm}(s)=\frac{s-1}{4}.
\end{equation}
For prompts that directly output a scalar score in $[0,1]$, such as Transition Naturalness and Narrative Coherence, the score is used without additional rescaling.

Aggregation is performed only within the valid evaluation scope of each metric. Frame-level metrics are averaged over sampled frames, shot-level metrics are averaged over matched shots, and transition-level metrics are averaged over detected shot boundaries:
\begin{equation}
S_m(V)
=
\frac{1}{|\mathcal{U}_m(V)|}
\sum_{u\in\mathcal{U}_m(V)}
s_m(u),
\end{equation}
where $m$ denotes a metric, $\mathcal{U}_m(V)$ denotes the valid evaluation units for video $V$, and $s_m(u)$ is the normalized score of each unit.

The final score of a model on each metric is computed by averaging over all applicable benchmark samples:
\begin{equation}
\bar{S}_m
=
\frac{1}{|\mathcal{D}_m|}
\sum_{V\in\mathcal{D}_m}
S_m(V),
\end{equation}
where $\mathcal{D}_m$ is the valid subset for metric $m$. For conditional metrics, such as Text Fidelity, Key-Region Similarity, Cross-shot Product Consistency, Transition Naturalness, and Audio Alignment, samples without the required evaluation targets are excluded rather than assigned zero scores. In particular, when a planned multi-shot case fails to produce multiple detected shots, the corresponding cross-shot metric is assigned a score of zero.

\subsection{Metric Validation}
\label{sup:metric_val}
We validate the GPT-assisted metrics of AdSpark-Bench by measuring their agreement with human rankings. Specifically, we focus on the six metrics that involve GPT-based judgment: Shot Execution Alignment, Content Alignment, Transition Naturalness, Advertisement Attractiveness, Creative Quality, and Narrative Coherence. These metrics cover instruction following, selling-point realization, transition quality, commercial appeal, creative quality, and multi-shot storytelling. To conduct this validation, we construct an additional evaluation set outside AdSpark-Bench, consisting of 100 held-out products. For each product, all evaluated models are used to generate advertisement videos under the same product reference and advertisement prompt. GPT-5.5 then assigns a score to each generated video along the six GPT-assisted dimensions according to the evaluation protocols defined in AdSpark-Bench. 
In parallel, we recruit six human raters with research backgrounds in video generation and advertisement-related research to independently evaluate the same generated videos using an aligned six-dimensional rubric. Each video is rated by multiple raters, and all raters are blinded to the identity of the generation model and the corresponding GPT-5.5 scores.

For each product and each metric dimension, we rank the generated videos from different models according to the aggregated human scores and the corresponding AdSpark-Bench metric scores, respectively. Let $P$ denote the number of products, $K$ = 19 the number of evaluated models, and $D$ the set of six GPT-assisted metric dimensions. For product $p$, model $k$, and dimension $d \in D$, we denote the averaged human score as $h_{p,k}^{d}$ and the corresponding AdSpark-Bench metric score as $m_{p,k}^{d}$. We convert these scores into rankings within each product group:
\[
r_{p,k}^{d,h} = \mathrm{rank}(h_{p,k}^{d}), \quad
r_{p,k}^{d,m} = \mathrm{rank}(m_{p,k}^{d}),
\]
where rank 1 indicates the best video among the $K$ model outputs for the same product. Tied scores are assigned average ranks. For each product $p$ and dimension $d$, we compute Spearman's rank correlation between the human ranking and the metric ranking:
\[
\rho_{p}^{d}
=
\mathrm{Spearman}
\left(
\{r_{p,k}^{d,h}\}_{k=1}^{K},
\{r_{p,k}^{d,m}\}_{k=1}^{K}
\right).
\]
When there are no ties, this is equivalent to:
\[
\rho_{p}^{d}
=
1 -
\frac{
6\sum_{k=1}^{K}
\left(r_{p,k}^{d,h} - r_{p,k}^{d,m}\right)^2
}{
K(K^2-1)
}.
\]
Finally, we average the correlations over all products to obtain the validation score for each dimension:
\[
\rho^{d}
=
\frac{1}{P}
\sum_{p=1}^{P}
\rho_{p}^{d}.
\]

We report $\rho^{d}$ separately for each of the six GPT-assisted dimensions in Tab.~\ref{tab:metric_validation}. Higher Spearman correlation indicates stronger alignment between the GPT-assisted metric and human preference. The GPT-assisted metrics show consistent positive correlations with human rankings across all dimensions. The agreement is stronger for relatively objective criteria such as shot execution and content alignment, while more subjective criteria such as creative quality and advertisement attractiveness exhibit lower but still meaningful correlations.

\input{Tables/sup_metric_validation}

\subsection{Benchmark Examples}

We present representative reference images from AdSpark-Bench in Figure~\ref{fig:benchmark_product}, with one example shown for each of its 40 product categories. These examples demonstrate the broad category coverage of the benchmark, spanning home appliances, digital electronics, clothing, furniture, personal care, food, industrial supplies, and other product domains.
We further provide a complete annotation example for the digital-electronics sample in Figure~\ref{sup:annotation_example_adspark_bench}. The example specifies the camera's visual identity and selling points, together with the corresponding scene and style designs, shot-level creative plans, aligned audio scripts, and generation prompts. It demonstrates how each benchmark case provides structured conditions for systematically evaluating product fidelity, instruction adherence, temporal coherence, audio alignment, and advertisement effectiveness.

\section{Additional Experimental Details}
\label{sup:detail_experiment}

\subsection{Details of Evaluation Models}

We evaluate representative reference-to-video generation models on AdSpark-Bench, covering both proprietary and open-source models. 

\paragraph{Proprietary Models.}
We include seven proprietary video generation systems. 
\textbf{ViduQ2} and \textbf{ViduQ3} are reference-conditioned video generation models from the Vidu series, supporting image-guided generation and multi-reference conditioning. ViduQ3 further supports native audio-video generation and places greater emphasis on complex motion and cinematic storytelling.
\textbf{Seedance 2.0}~\cite{seedance2026seedance} is a unified multimodal audio-video generation model that accepts text, image, video, and audio references, enabling reference-guided generation with native audio.
\textbf{HappyHorse-1.1} is a joint audio-video generation model supporting text-to-video, image-to-video, and reference-to-video generation.
\textbf{PixVerse V5} and \textbf{PixVerse V6} are general-purpose video generation systems supporting image-conditioned synthesis, while the latter additionally provides more flexible reference-based generation and video extension capabilities.
\textbf{Kling 3.0 Omni}~\cite{team2025kling} is a multimodal video generation system that unifies reference-guided generation, video editing, and visual-language instruction following.

\paragraph{Open-Source Models.}
We further evaluate eleven open-source models with different reference-conditioning mechanisms.
\textbf{VACE-14B}~\cite{jiang2025vace} unifies reference-to-video generation and multiple video editing tasks through a shared video-conditioning interface.
\textbf{SkyReels-V3-14B}~\cite{li2026skyreels} adopts a multimodal in-context generation framework and supports reference-image-to-video generation, video extension, and audio-guided synthesis.
\textbf{Phantom-14B}~\cite{liu2025phantom} performs single- and multi-subject video generation through joint text--image conditioning and cross-modal alignment.
\textbf{VINO-13B}~\cite{chen2026vino} is a unified visual generation model that handles interleaved multimodal conditions for controllable image and video synthesis.
\textbf{RefAlign-14B}~\cite{wang2026refalign} explicitly aligns reference-branch representations with visual foundation model features to improve reference fidelity and reduce subject confusion.
\textbf{Kaleido-14B}~\cite{zhang2025kaleido} targets multi-subject reference video generation and introduces reference-aware positional encoding for stable multi-image conditioning.
\textbf{HunyuanCustom-13B}~\cite{hu2025hunyuancustom} is a customized video generation framework that integrates multimodal reference information to preserve target subject characteristics.
\textbf{Bernini-14B}~\cite{team2026bernini} employs latent semantic planning to improve compositional organization and semantic control during video generation.
\textbf{MV-S2V-14B}~\cite{song2026mv} conditions generation on multiple views of a target subject to improve three-dimensional subject consistency.
\textbf{MAGREF-14B}~\cite{deng2025magref} introduces masked reference guidance for flexible any-reference video generation.
Finally, \textbf{LTX-2-14B}~\cite{hacohen2026ltx} is an efficient joint audio-video foundation model supporting synchronized visual and audio generation.

\paragraph{Evaluation Setup.}
For fair comparison, all models are evaluated using the same product reference images and advertisement prompts from AdSpark-Bench whenever their interfaces support reference-conditioned generation. We follow the official inference settings of each model. For models without audio generation capability, audio-related metrics are marked as unavailable rather than assigned zero scores.

\subsection{Implementation Details}

We fine-tune LTX-2 as the AdSpark-adapted baseline, denoted as LTX-AdSpark, to validate the effectiveness of AdSpark-300K. We use LTX-2 as the base model and adopt LoRA training instead of full-parameter fine-tuning for efficient adaptation.

\paragraph{Training Configuration.}
LoRA is applied to the attention projection layers, including \texttt{to\_q}, \texttt{to\_k}, and \texttt{to\_v}. The LoRA rank and alpha are both set to 64, with dropout set to 0. The model is trained for 30K steps using AdamW with a learning rate of $2\times10^{-4}$, batch size 4, gradient clipping of 1.0, and bfloat16 mixed precision. Gradient checkpointing is enabled during training.

\paragraph{Conditioning Strategy.}
During training and inference, the model uses the product reference image as the primary visual condition and the structured advertisement prompt as textual guidance. Video and audio latents are precomputed from AdSpark-300K, while audio generation is learned from text supervision without reference audio conditioning.

\paragraph{Inference Setting.}
For validation and benchmark inference, we generate 9:16 advertisement videos at $704\times1280$ resolution with 121 frames at 24 fps, corresponding to approximately 5 seconds. We use 30 inference steps, guidance scale 4.0, STG scale 1.0, and a fixed random seed of 42. Checkpoints are saved every 1,000 steps, and validation is performed every 3,000 steps.

\subsection{Qualitative Comparisons}
\label{sup:case}
We provide additional qualitative comparisons on AdSpark-Bench in Fig.~\ref{fig:case1}-Fig.\ref{fig:case3}. Each example uses the same product reference image and structured advertisement prompt across all models. The results show that existing models can generally synthesize plausible video, but they still struggle with fine-grained product preservation, interaction realism, and shot-level prompt following. For instance, several methods change the geometry or material details of the clothes-drying rack, distort the drawer structure of the wooden desk, or fail to preserve the bracelet shape and pendant details in close-up shots. Some models also introduce inconsistent hands, abrupt viewpoint changes, or weak correspondence between the planned demonstration and the generated motion. In comparison, LTX-AdSpark produces more product-consistent results with smoother shot progression and clearer selling-point presentation, indicating that AdSpark-300K improves reference-conditioned advertisement video generation in product-centric scenarios.

\subsection{Ablation Results}
\input{Tables/ablation_data_composition}

To disentangle the effects of data composition and dataset scale, we further fine-tune LTX-2 on different subsets of AdSpark-300K under identical optimization settings. As shown in Tab.~\ref{tab:data_composition_ablation}, both real-world and synthetic advertisements consistently improve over the base model, while exhibiting complementary strengths.
We first compare \textit{Real-100K} and \textit{Synthetic-100K} under the same training scale, where \textit{Synthetic-100K} is randomly sampled from the full 200K synthetic subset. The two subsets show different strengths. Real-world data performs better on Subject Consistency (64.73 vs.\ 58.17), suggesting that authentic product appearances provide stronger identity supervision. Synthetic data is more effective on structure- and transition-related metrics, reaching 71.79 on Shot Structure Alignment, 80.09 on Transition Naturalness, and 81.38 on Narration Script Consistency, compared with 70.68, 72.76, and 78.26 for Real-100K. This is consistent with its construction from explicit shot-level plans and aligned narration instructions.
Scaling the synthetic subset from 100K to 200K brings further gains across all reported metrics, showing that data scale also contributes. Nevertheless, the full AdSpark-300K performs best overall. In particular, adding 100K real-world samples on top of Synthetic-200K raises Subject Consistency from 61.59 to 65.01 and further improves the remaining metrics. These results suggest that the final performance gains come from both increased training scale and the complementary supervision of the real-world and synthetic subsets.

\section{Broader Impact and Limitations}

\subsection{Limitations and Future Work}

AdSpark provides an important step toward product-centric advertisement video generation, while leaving several directions for future extension. First, the current dataset is constructed from e-commerce scenarios. Future work could extend the data sources to more diverse platforms, markets, and advertising styles. Second, our product selection is mainly guided by click-through rate, which helps prioritize representative and commercially relevant products with strong user interest. At the same time, future extensions could further cover a broader long-tail product distribution, including niche categories and less frequently promoted products, to support more comprehensive evaluation and generation across diverse product types.

\input{Prompt/vlm_quality_filter}
\input{Prompt/real_world_ad_ann}
\input{Prompt/unified_ann}
\input{Prompt/synthetic_planning_prompt}
\input{Prompt/shot_camera_motion}
\input{Prompt/synthetic_plan_review}
\input{Prompt/ann_example_adspark_300k}
\input{Prompt/ann_example_adspark-bench}
\input{Prompt/shot_execution_alignment}
\input{Prompt/scene_style_alignment}
\input{Prompt/selling_point_realization}
\input{Prompt/transition_naturalness}
\input{Prompt/advertisement_attractive}
\input{Prompt/creative_quality}
\input{Prompt/narr_coherence}

%% file: Tables/sup_three_stage_filter.tex
\begin{table*}[t]
\centering
\small
\label{tab:synthetic_sku_filtering}
\setlength{\tabcolsep}{5pt}
\begin{tabular}{lrrr}
\toprule
\textbf{Stage} & \textbf{Input} & \textbf{Removed} & \textbf{Retained} \\
\midrule
Quality Filtering     & 600,000 & 199,870 & 400,130 \\
Suitability Filtering & 400,130 & 49,290  & 350,840 \\
Category Filtering    & 350,840 & 2,314   & 348,526 \\
\bottomrule
\end{tabular}
\caption{\textbf{Statistics of the three-stage SKU filtering pipeline.}
The number of removed SKUs is computed relative to the preceding stage.}
\end{table*}

%% file: tex/statistics_supplementary.tex
\begin{figure*}
    \centering
    \includegraphics[width=\linewidth]{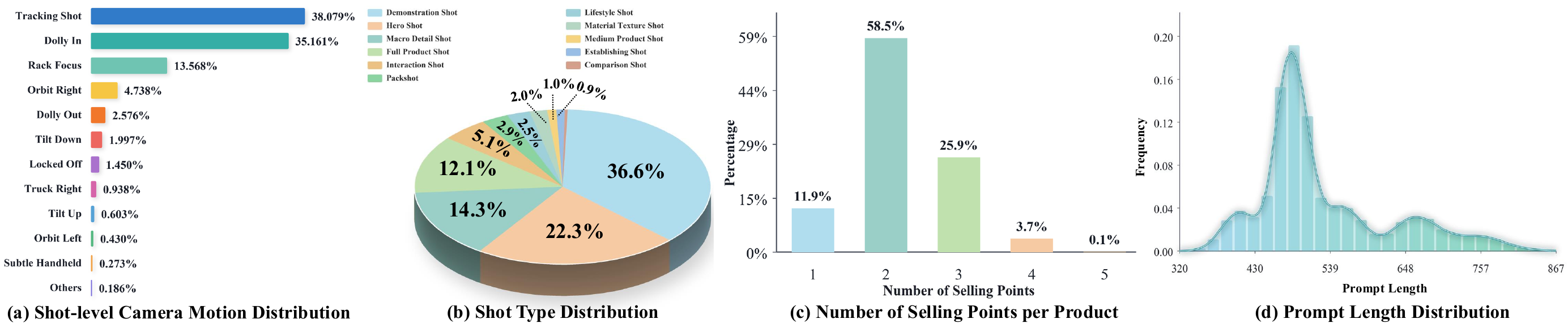}
    \caption{\textbf{Additional statistics of AdSpark-300K.}
The figure summarizes the distributions of (a) shot-level camera motions, (b) shot types, (c) the number of selling points per product, and (d) advertisement prompt lengths, illustrating the diversity of camera design, product presentation, commercial content, and prompt complexity
in the dataset.}
\label{fig:additional_dataset_statistics}
\end{figure*}

%% file: tex/benchmark_product.tex
\begin{figure*}[t]
    \centering
    \includegraphics[width=\linewidth]{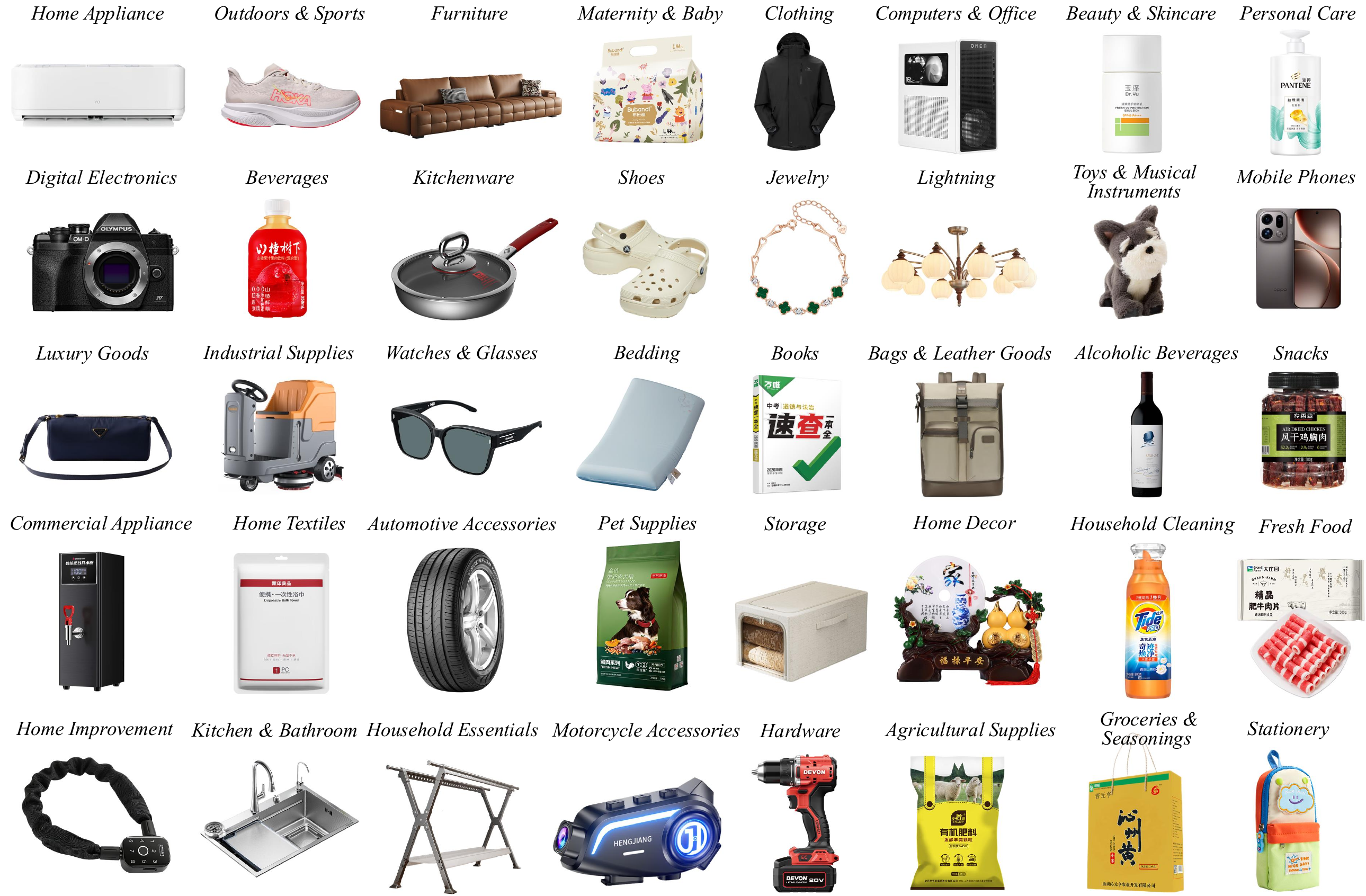}
    \caption{\textbf{Representative reference images from AdSpark-Bench.} We show benchmark reference products across diverse categories, including home appliances, electronics, clothing, furniture, food, personal care, and industrial supplies.}
    \label{fig:benchmark_product}
\end{figure*}

%% file: tex/dataset_presentation.tex
\begin{figure*}[!t]
    \centering
    \includegraphics[width=\linewidth]{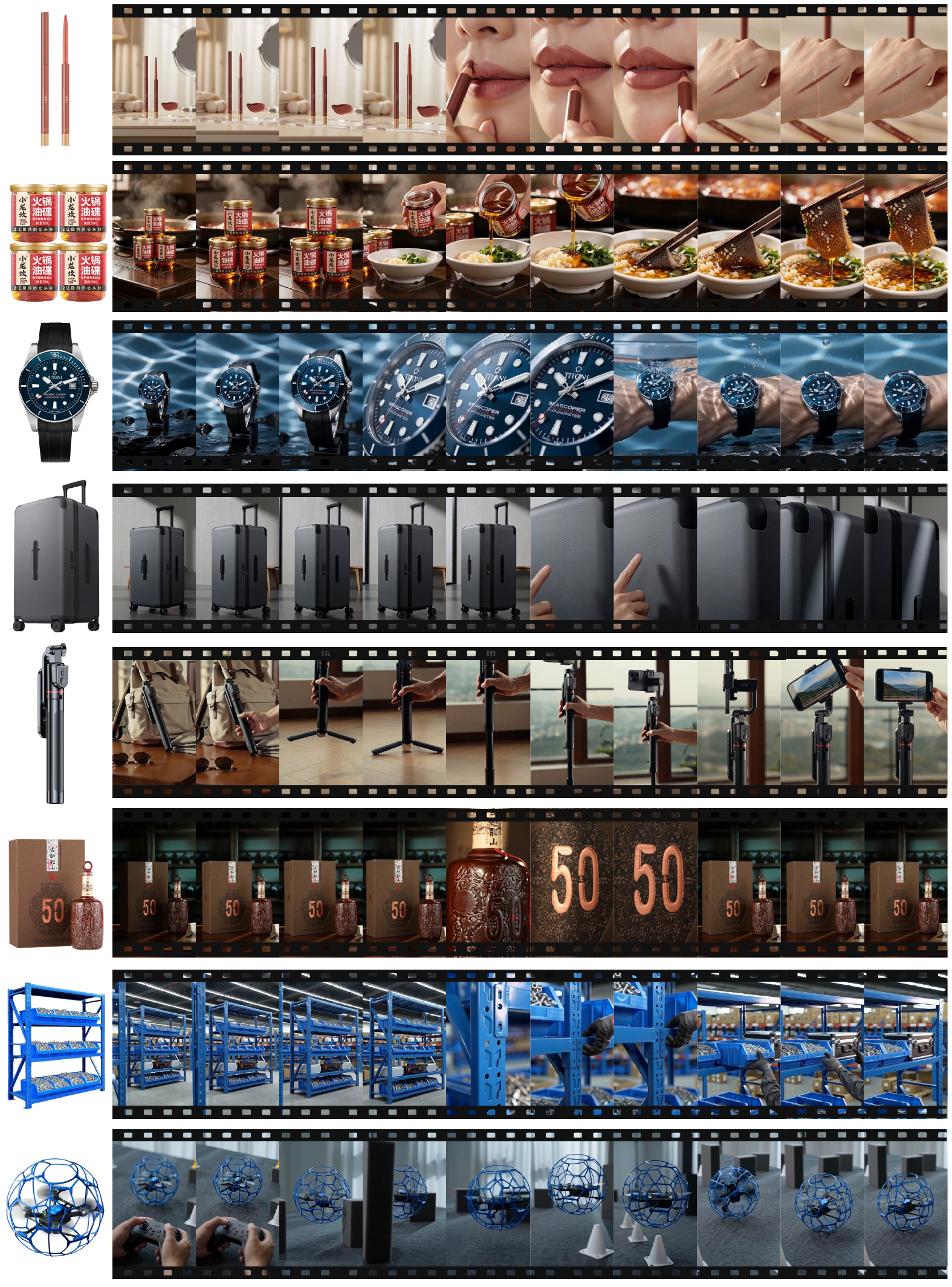}
    \caption{\textbf{Qualitative examples from AdSpark-300K.} 
Each row presents a reference product image and paired advertisement video frames sampled in temporal order.}
    \label{fig:dataset_presentation}
\end{figure*}

%% file: tex/annotation.tex
\begin{figure*}[!t]
    \centering
    \includegraphics[width=\linewidth]{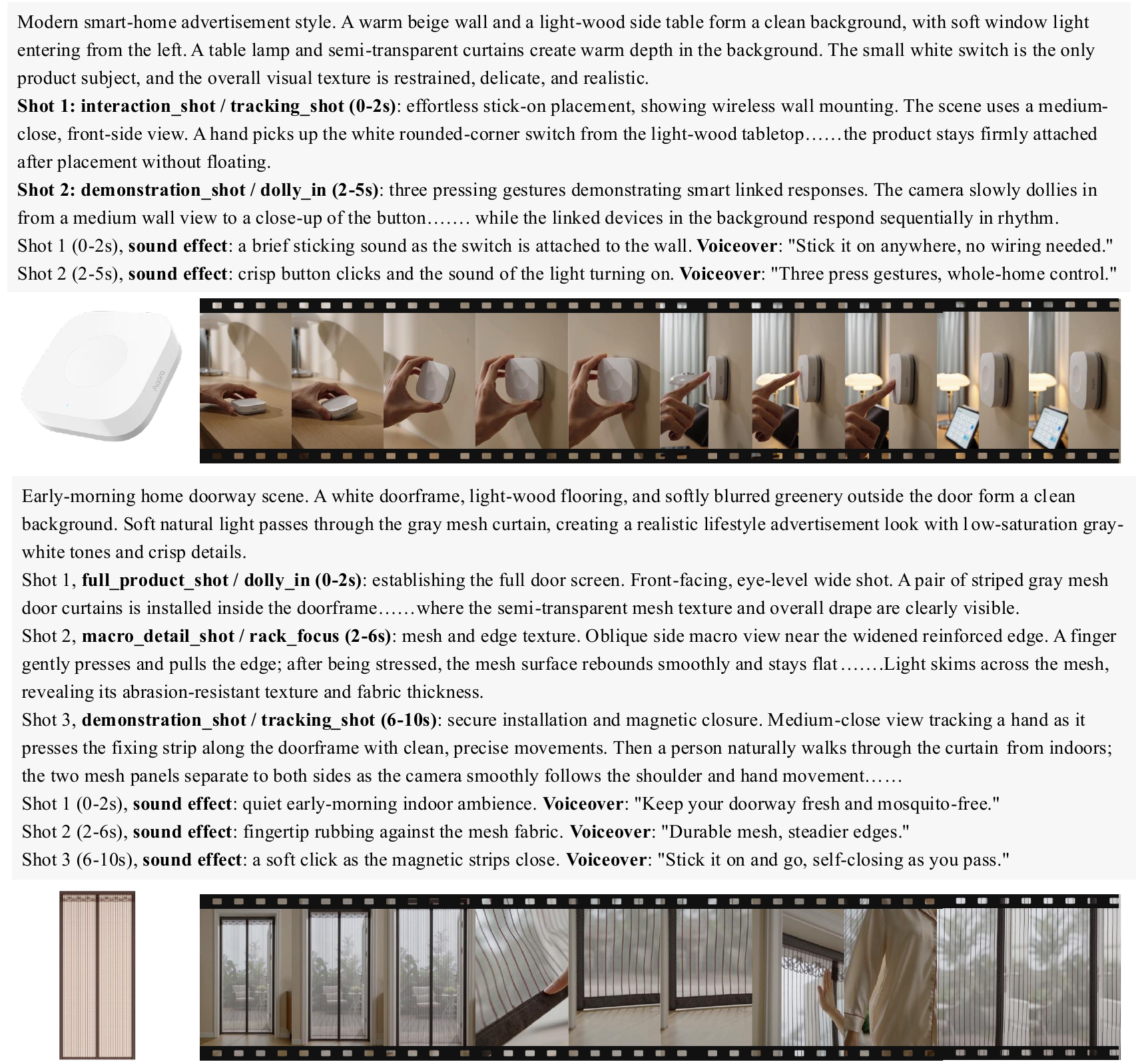}
    \caption{\textbf{Examples of advertisement videos and their structured prompts in AdSpark-300K.} For each product, the prompt describes the overall scene style, shot-level visual planning, camera type and motion, temporal segments, and synchronized audio cues, including sound effects and voiceover scripts.
}
    \label{fig:example_annotatio_video}
\end{figure*} 

%% file: Algorithm/alg_product_fidelity.tex
\begin{algorithm}[t]
\caption{Product Fidelity Evaluation Protocol}
\label{alg:product_fidelity}
\small
\begin{algorithmic}[1]
\STATE \textbf{Input:} Reference image \(I_r\), generated video \(V\), product annotations \(\mathcal{A}\).
\STATE \textbf{Output:} Individual product fidelity metrics.
\STATE Uniformly sample frames from \(V\): \(\displaystyle \mathcal{F}=\{f_i\}_{i=1}^{N}\).
\STATE Detect product regions in \(I_r\) and \(\mathcal{F}\) using GroundingDINO.
\STATE Refine detected regions with SAM2: \(\displaystyle \mathcal{P}^{r},\ \mathcal{P}^{g}=\{P_i^{g}\}_{i=1}^{N}\).
\STATE Encode product crops with DINOv3: \(\displaystyle z_r=\phi(\mathcal{P}^{r}),\ z_i=\phi(P_i^{g})\).
\STATE Compute Subject Consistency: \(\displaystyle S_{\mathrm{subj}}\leftarrow\frac{1}{N}\sum_{i=1}^{N}\cos(z_r,z_i)\).
\STATE Localize annotated identity regions, including logos, brand marks, and packaging details.
\STATE Compute Key-Region Similarity: \(\displaystyle S_{\mathrm{region}}\leftarrow\frac{1}{|\mathcal{R}|}\sum_{j\in\mathcal{R}}\cos(z_j^r,z_j^g)\).
\STATE Apply OCR to localized product crops and obtain recognized text \(\hat{y}\).
\STATE Compute Text Fidelity: \(\displaystyle S_{\mathrm{text}}\leftarrow 1-\frac{d_{\mathrm{edit}}(y,\hat{y})}{|y|}\).
\STATE Detect shot boundaries with TransNetV2 for multi-shot videos.
\STATE Compute Cross-shot Product Consistency when transitions exist: \(\displaystyle S_{\mathrm{xshot}}\leftarrow\frac{1}{K}\sum_{k=1}^{K}\cos(z_k^{-},z_k^{+})\).
\STATE Return all applicable product fidelity metrics independently.
\end{algorithmic}
\end{algorithm}

%% file: Algorithm/alg_instruction_adherence.tex
\begin{algorithm}[t]
\caption{Instruction Adherence Evaluation Protocol}
\label{alg:instruction_adherence}
\small
\begin{algorithmic}[1]
\STATE \textbf{Input:} Generated video \(V\), creative plan \(\mathcal{C}\), advertisement prompt \(p\).
\STATE \textbf{Output:} Individual instruction adherence metrics.
\STATE Detect generated shots and parse planned shots from \(\mathcal{C}\).
\STATE Match generated shots to planned shots using temporal IoU.
\STATE Compute shot-count accuracy:
\(\displaystyle S_{\mathrm{count}} \leftarrow \max\left(0,1-\frac{|N_g-N_p|}{N_p}\right).\)
\STATE Compute boundary accuracy:
\(\displaystyle S_{\mathrm{bound}} \leftarrow \frac{1}{|\mathcal{B}^{p}|}\sum_{\tau \in \mathcal{B}^{p}}
\mathbf{1}\!\left[\min_{b\in\mathcal{B}^{g}}|\tau-b|\leq \delta\right].\)
\STATE Obtain Shot Structure Alignment from \(S_{\mathrm{count}}\) and \(S_{\mathrm{bound}}\).
\STATE Construct contact sheets for matched shots and estimate shot execution with GPT-5.5:
\(\displaystyle S_{\mathrm{exec}} \leftarrow \frac{1}{|\mathcal{M}|}
\sum_{(i,j)\in\mathcal{M}} q^{\mathrm{exec}}_{ij}.\)
\STATE Evaluate scene, style, and selling-point realization with GPT-5.5:
\(\displaystyle S_{\mathrm{content}} \leftarrow \frac{1}{|\mathcal{M}|}
\sum_{(i,j)\in\mathcal{M}} q^{\mathrm{content}}_{ij}.\)
\STATE Encode the prompt and sampled frames using GME:
\(\displaystyle S_{\mathrm{gme}} \leftarrow \frac{1}{N}\sum_{i=1}^{N}
\psi_t(p)^\top \psi_v(f_i).\)
\STATE Return all applicable instruction adherence metrics independently.
\end{algorithmic}
\end{algorithm}

%% file: Algorithm/alg_temporal_coherence.tex
\begin{algorithm}[t]
\caption{Temporal Coherence Evaluation Protocol}
\label{alg:temporal_coherence}
\small
\begin{algorithmic}[1]
\STATE \textbf{Input:} Generated video \(V\), detected shot sequence \(\mathcal{S}^{g}\).
\STATE \textbf{Output:} Individual temporal coherence metrics.
\STATE For each detected shot, compute optical flow between adjacent frames.
\STATE Obtain frame-wise motion magnitudes:
\STATE \(\displaystyle m_t \leftarrow \frac{1}{|\Omega|}\sum_{x\in\Omega}\|u_t(x)\|_2.\)
\STATE Compute motion amplitude:
\STATE \(\displaystyle A \leftarrow \frac{1}{T-1}\sum_{t=1}^{T-1}m_t.\)
\STATE Compute motion smoothness:
\STATE \(\displaystyle S_{\mathrm{smooth}} \leftarrow \max\left(0,1-\frac{\sigma(m)}{\mu(m)+\epsilon}\right).\)
\STATE Track product masks within each shot using GroundingDINO and SAM2.
\STATE Compute product motion stability:
\STATE \(\displaystyle S_{\mathrm{stab}} \leftarrow 0.4S_{\mathrm{cent}}+0.3S_{\mathrm{area}}+0.3S_{\mathrm{iou}}.\)
\STATE Detect all shot boundaries.
\STATE Estimate Transition Naturalness around each boundary:
\STATE \(\displaystyle S_{\mathrm{trans}} \leftarrow \frac{1}{K}\sum_{k=1}^{K} q_k^{\mathrm{trans}}.\)
\STATE Detect repeated, flickering, and corrupted frames around boundaries.
\STATE Compute Temporal Consistency:
\STATE \(\displaystyle S_{\mathrm{temp}} \leftarrow \frac{1}{K}\sum_{k=1}^{K} q_k^{\mathrm{temp}}.\)
\STATE Return all applicable temporal coherence metrics independently.
\end{algorithmic}
\end{algorithm}

%% file: Algorithm/alg_audio_alignment.tex
\begin{algorithm}[t]
\caption{Audio Alignment Evaluation Protocol}
\label{alg:audio_alignment}
\small
\begin{algorithmic}[1]
\STATE \textbf{Input:} Generated video \(V\), generated audio \(a\), audio description \(d_a\), reference script \(y\).
\STATE \textbf{Output:} Individual audio alignment metrics.
\STATE Extract the audio track \(a\) from the generated video.
\STATE Encode \(a\) and \(d_a\) using LAION-CLAP.
\STATE Compute Background Sound Consistency:
\(\displaystyle S_{\mathrm{bg}}\leftarrow \cos(\eta_a(a),\eta_t(d_a))\).
\STATE Transcribe narration using Whisper-Large-v3-Turbo to obtain \(\hat{y}\).
\STATE Compute Narration Script Consistency:
\(\displaystyle S_{\mathrm{script}}\leftarrow 1-\frac{d_{\mathrm{edit}}(y,\hat{y})}{|y|}\).
\STATE Return all applicable audio alignment metrics independently.
\end{algorithmic}
\end{algorithm}

%% file: Algorithm/alg_ad_effectiveness.tex
\begin{algorithm}[t]
\caption{Advertisement Effectiveness Evaluation}
\label{alg:ad_effectiveness}
\small
\begin{algorithmic}[1]
\STATE \textbf{Input:} Reference image \(I_r\), generated video \(V\), creative plan \(\mathcal{C}\), selling points \(\mathcal{P}\).
\STATE \textbf{Output:} Individual advertisement effectiveness metrics.
\STATE Uniformly sample representative frames from \(V\).
\STATE Construct a visual summary using sampled frames and shot-level annotations.
\STATE Provide GPT-5.5 with \(I_r\), \(\mathcal{C}\), \(\mathcal{P}\), and the visual summary.
\STATE Evaluate Advertisement Attractiveness:
\(\displaystyle S_{\mathrm{attr}}\leftarrow \mathrm{Norm}(q_{\mathrm{attr}})\).
\STATE Evaluate Creative Quality:
\(\displaystyle S_{\mathrm{creat}}\leftarrow \mathrm{Norm}(q_{\mathrm{creat}})\).
\STATE Evaluate Narrative Coherence:
\(\displaystyle S_{\mathrm{narr}}\leftarrow \mathrm{Norm}(q_{\mathrm{narr}})\).
\STATE Return all advertisement effectiveness metrics independently.
\end{algorithmic}
\end{algorithm}

%% file: tex/case1.tex
\begin{figure*}[!t]
    \centering
    \includegraphics[width=\linewidth]{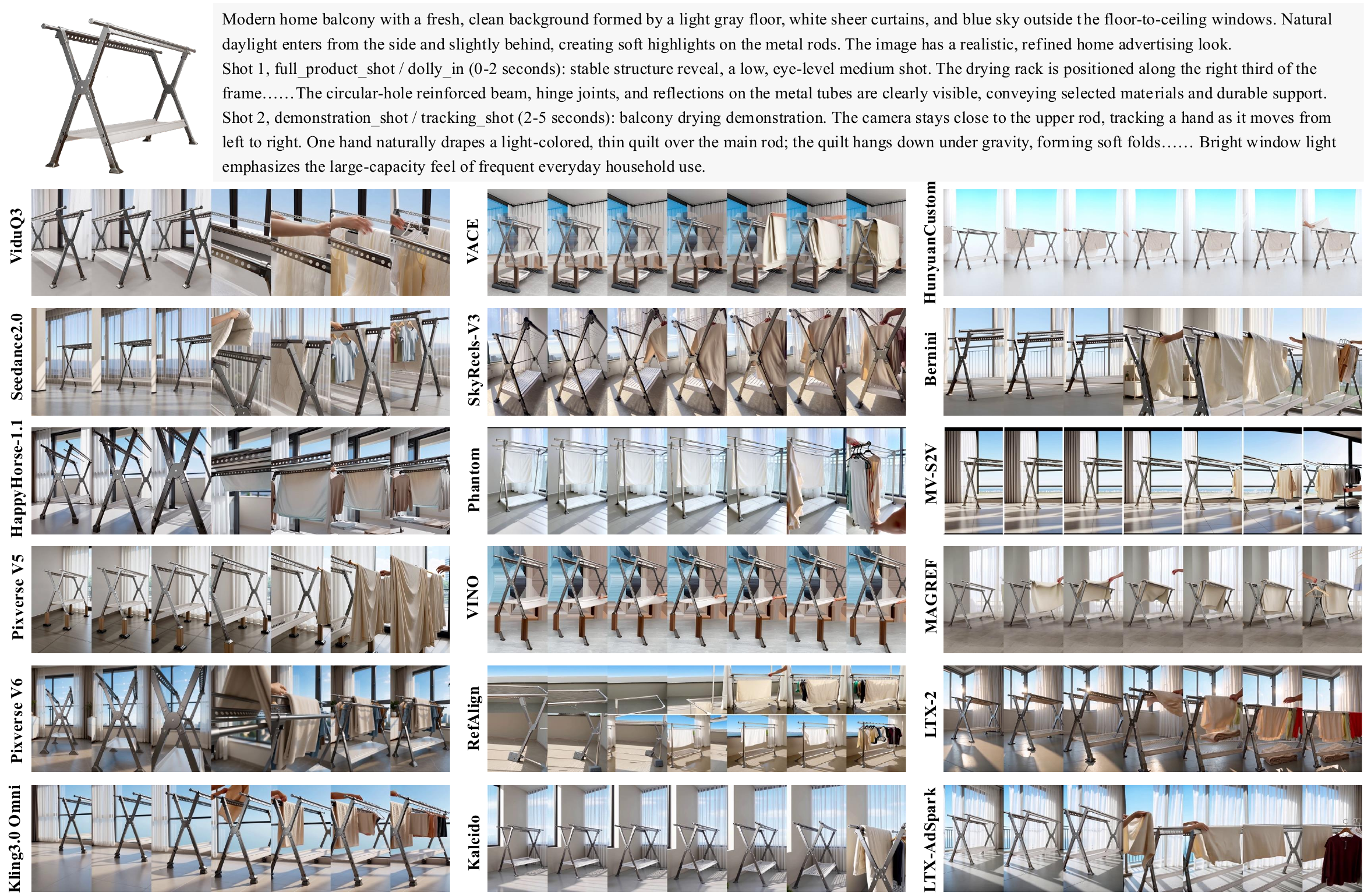}
    \caption{Qualitative comparison of different models for a clothes-drying rack product.}
\label{fig:case1}
\end{figure*}

%% file: tex/case2.tex
\begin{figure*}[!t]
    \centering
    \includegraphics[width=\linewidth]{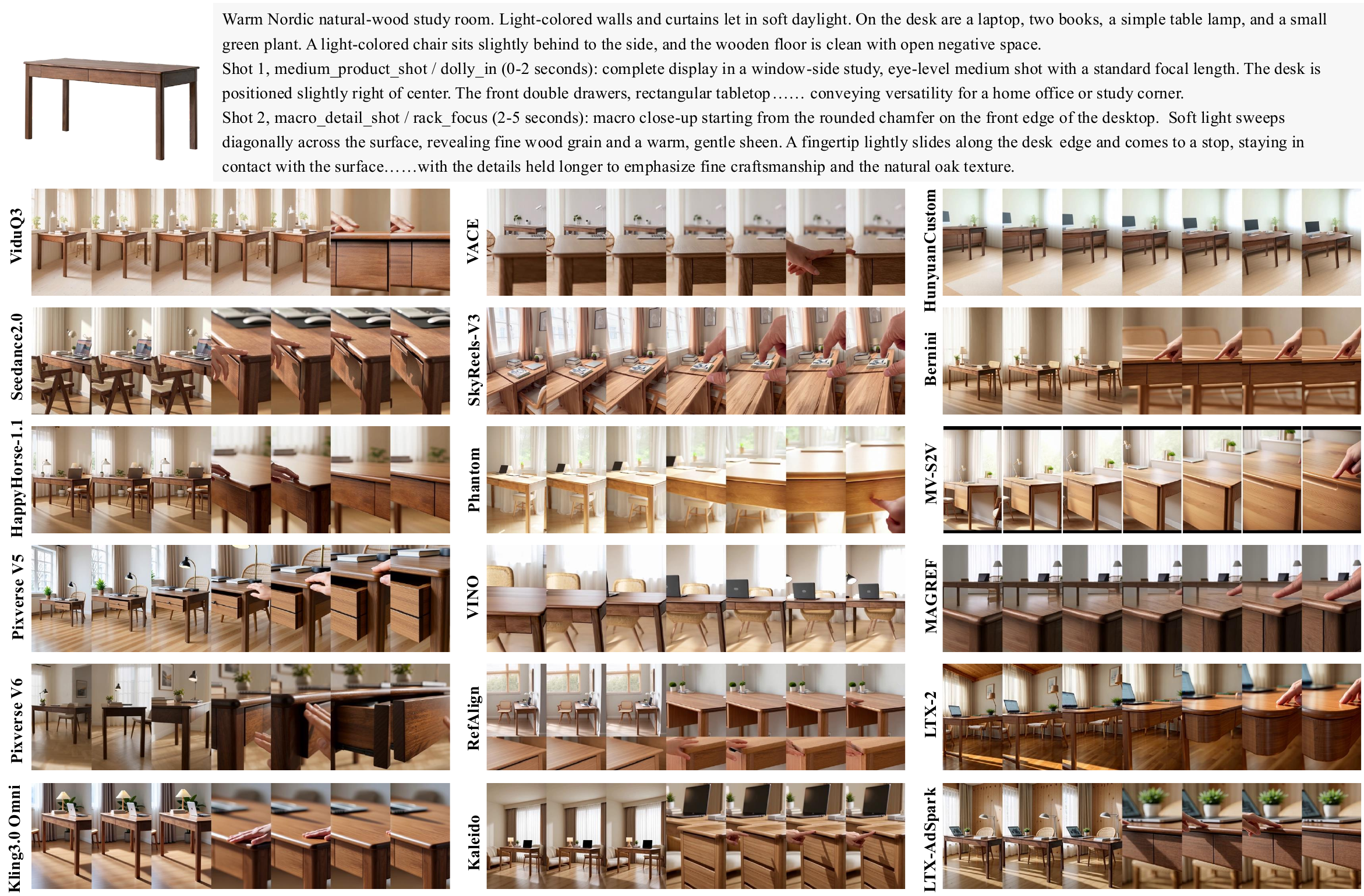}
    \caption{Qualitative comparison of different models for a wooden desk product.}
\label{fig:case2}
\end{figure*}

%% file: tex/case3.tex
\begin{figure*}[!t]
    \centering
    \includegraphics[width=\linewidth]{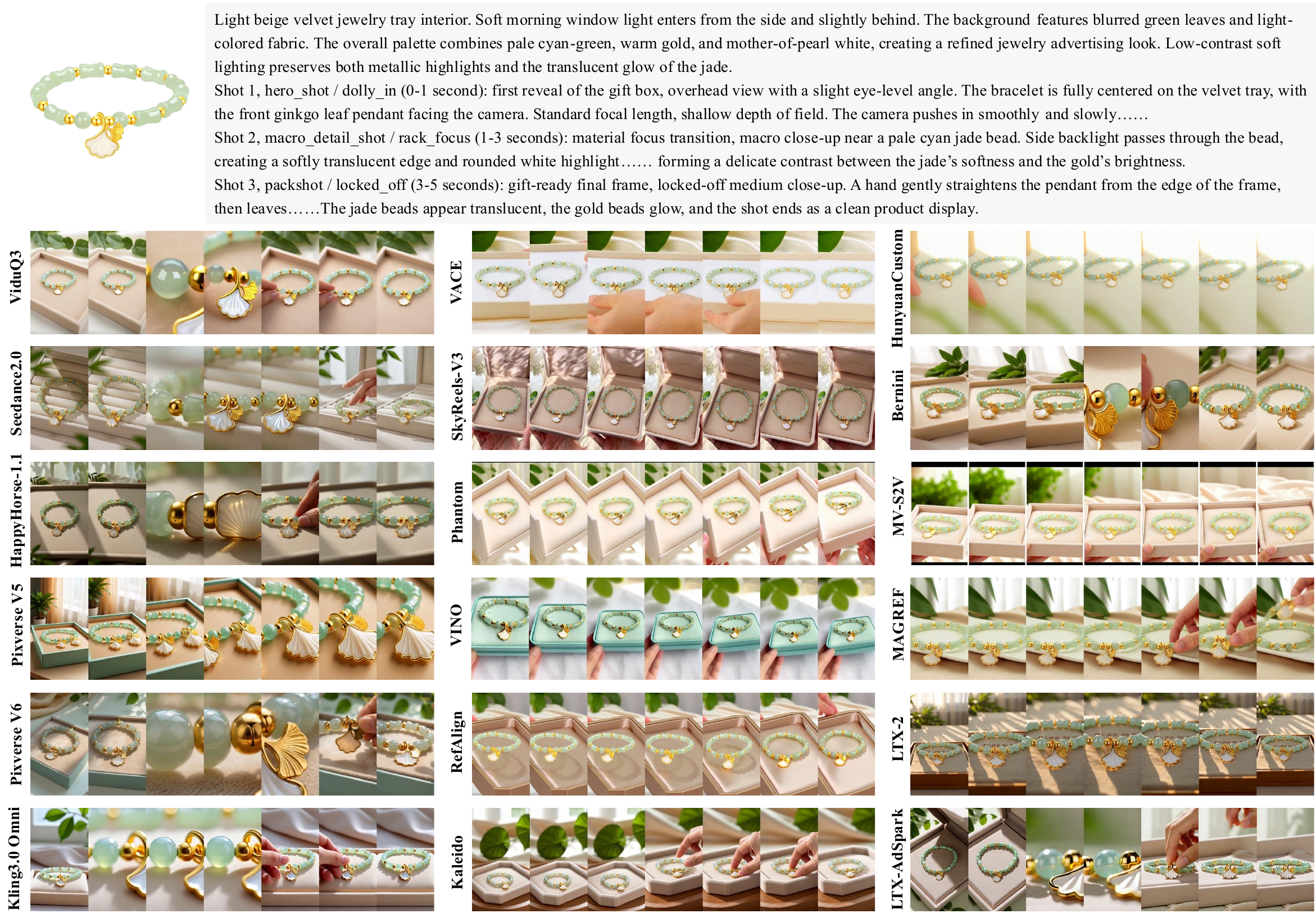}
    \caption{Qualitative comparison of different models for a jewelry product.}
\label{fig:case3}
\end{figure*}

%% file: Tables/sup_metric_validation.tex
\begin{table*}[t]
\centering
\begin{tabular}{lc}
\toprule
Metric & Spearman $\rho$ $\uparrow$ \\
\midrule
Shot Execution Alignment & 0.861 \\
Content Alignment & 0.833 \\
Transition Naturalness & 0.797 \\
Advertisement Attractiveness & 0.719 \\
Creative Quality & 0.752 \\
Narrative Coherence & 0.774 \\
\bottomrule
\end{tabular}
\caption{Correlation between GPT-assisted metrics and human rankings.}
\label{tab:metric_validation}
\end{table*}

%% file: Tables/ablation_data_composition.tex
\begin{table*}[t]
\centering
\caption{
Ablation on the composition and scale of AdSpark-300K. All variants are initialized from the same LTX-2 checkpoint and trained with
identical optimization settings unless otherwise specified. The best results are highlighted in \textbf{bold}.
}
\label{tab:data_composition_ablation}
\resizebox{\textwidth}{!}{
\begin{tabular}{lcc|cccccc}
\toprule
\textbf{Training Data} &
\textbf{Real} &
\textbf{Synthetic} &
\textbf{Img.} &
\textbf{Subj.} &
\textbf{Struct.} &
\textbf{Trans.} &
\textbf{Script} &
\textbf{Creat.} \\
\midrule

LTX-2 (Base)
& -- & --
& 66.17
& 50.39
& 46.29
& 33.66
& 74.71
& 56.05 \\

\midrule

Real-100K
& 100K & 0
& 68.03
& 64.73
& 70.68
& 72.76
& 78.26
& 57.61 \\

Synthetic-100K
& 0 & 100K
& 68.44
& 58.17
& 71.79
& 80.09
& 81.38
& 57.12 \\

Synthetic-200K
& 0 & 200K
& 68.69
& 61.59
& 76.23
& 87.26
& 82.62
& 58.47 \\

\midrule

AdSpark-300K
& 100K & 200K
& \textbf{69.57}
& \textbf{65.01}
& \textbf{77.53}
& \textbf{88.09}
& \textbf{84.41}
& \textbf{59.29} \\

\bottomrule
\end{tabular}
}
\end{table*}

%% file: Prompt/vlm_quality_filter.tex
\begin{figure*}[!t]
\centering
\begin{tcolorbox}[
    colback=gray!5,
    colframe=gray!60,
    title=\textbf{Prompt for VLM-based Quality Filtering},
    fonttitle=\bfseries,
    width=\textwidth,
    boxrule=0.5pt
]
\small

You are filtering a candidate video clip for constructing a high-quality product-centric advertisement video dataset.

\textbf{Input Organization}:  
Image 1 is the transparent-background product reference image. Images 2 to $N$ are temporally ordered frames sampled from the same contiguous candidate clip. The product caption and the detected product bounding box for each sampled frame are also provided.

\textbf{Evaluation Objective}:  
Determine whether the candidate clip is suitable for inclusion in the dataset. The clip should preserve the identity of the reference product and present it clearly throughout all sampled frames.

\textbf{Product Identity Consistency}:  
Verify that every sampled frame contains the same product as the reference image. Product identity should be assessed using fine-grained visual characteristics, including:

\begin{itemize}
    \item packaging appearance;
    \item brand logos and distinctive markings;
    \item dominant colors and materials;
    \item product shape, structure, and proportions.
\end{itemize}

If the product in any sampled frame differs from the reference image, set \texttt{matches\_reference\_product} to \texttt{false} and reject the clip.

\textbf{Product Visibility}:  
Reject the clip if the product in any sampled frame is:

\begin{itemize}
    \item only partially visible;
    \item severely occluded;
    \item too small to be reliably identified;
    \item incorrectly localized by the provided bounding box.
\end{itemize}

Set \texttt{is\_mostly\_visible} to \texttt{true} only when the product remains sufficiently complete and recognizable in all sampled frames.

\textbf{Overlay Artifact Detection}:  
Reject the clip if large floating text or graphics substantially obscure the product. Overlay artifacts include subtitles, captions, stickers, promotional labels, or marketing graphics rendered over the video frame. Text printed directly on the product or its packaging should not be treated as an overlay artifact.

\textbf{Decision Rule}:  
The candidate clip is suitable only when all sampled frames satisfy the product identity, visibility, and overlay-artifact requirements. If any sampled frame fails any requirement, set \texttt{suitable\_for\_gt} to \texttt{false}.

\textbf{Output Format}:  
Return only a valid JSON object following the schema below:

\begin{lstlisting}
{
    "suitable_for_gt": true,
    "matches_reference_product": true,
    "has_heavy_text_overlay": false,
    "is_mostly_visible": true,
    "reason": "The product matches the reference image and remains clearly visible in all sampled frames."
}
\end{lstlisting}

\textbf{Input Fields}:

Reference Product Image:

Sampled Video Frames:

Product Caption:

Per-frame Product Bounding Boxes:

\end{tcolorbox}
\caption{Prompt for VLM-based Quality Filtering.}
\label{sup:vlm_filtering_prompt}
\end{figure*}

%% file: Prompt/real_world_ad_ann.tex
\begin{figure*}[!t]
\centering
\begin{tcolorbox}[
    colback=gray!5,
    colframe=gray!60,
    title={\textbf{Prompt for Real-World Advertisement Annotation}},
    fonttitle=\bfseries,
    width=\textwidth,
    boxrule=0.5pt
]
\small

You are annotating a product-centric advertisement video. Image 1 is the transparent-background product reference image. The remaining images are representative frames sampled from temporally ordered shots of the same advertisement clip. Product metadata, detected shot boundaries, and the timestamped audio transcript are also provided.

\textbf{Task}:  
Generate structured annotations that accurately describe the advertised product and the content actually presented in the video. All annotations must be grounded in the provided reference image, video frames, product metadata, and transcript. Do not introduce unsupported product attributes, selling points, actions, scenes, or transitions.

\textbf{Product Identity}:  
Describe the product category, brand, product name, and identity-defining visual attributes. The visual identity should include dominant colors, materials, shape, key regions, recognizable logos or text, and essential details that should remain unchanged across the video. The description must be consistent with both the reference image and the sampled frames.

\textbf{Selling Points}:  
Identify the product features or benefits explicitly supported by the product metadata, narration, or visual content. For each selling point, describe how it is presented in the video and specify observable criteria indicating successful visualization. Do not introduce unsupported commercial claims.

\textbf{Creative Plan}:  
Summarize the scene and visual style observed in the video. For each detected shot, annotate its temporal range, shot type, camera motion, visible content, and the selling points it presents. Preserve the original shot order and do not add, merge, or invent shots.

\textbf{Audio Script}:  
Use the provided transcript and timestamps to associate narration and relevant background sounds with the corresponding shots. Preserve the original meaning of the narration and omit unrelated, unintelligible, or non-advertisement speech.

\textbf{Output Format}:  
Return only a valid JSON object following the schema below:

\begin{lstlisting}
Return only a valid JSON object following the schema shown in Fig. 14.
\end{lstlisting}

\textbf{Input Fields}:

Reference Product Image:

Temporally Ordered Shot Frames:

Product Metadata:

Detected Shot Boundaries:

Timestamped Audio Transcript:

\end{tcolorbox}
\caption{Prompt for real-world advertisement annotation.}
\label{sup:realworld_annotation_prompt}
\end{figure*}

%% file: Prompt/unified_ann.tex
\begin{figure*}[!t]
\centering
\begin{tcolorbox}[
    colback=gray!5,
    colframe=gray!60,
    title={\textbf{Unified Annotation Schema of AdSpark-300K}},
    fonttitle=\bfseries,
    width=\textwidth,
    boxrule=0.5pt
]
\footnotesize

\begin{lstlisting}[
    basicstyle=\ttfamily\scriptsize,
    columns=fullflexible,
    keepspaces=true,
    lineskip=-1pt,
    aboveskip=2pt,
    belowskip=2pt
]
{
  "product_identity": {
    "category": "",
    "brand": "",
    "product_name": "",
    "visual_identity": {
      "main_colors": [],
      "materials": [],
      "shape": "",
      "key_regions": [],
      "ocr_targets": [],
      "must_keep": []
    }
  },
  "selling_points": [
    {
      "point": "",
      "visualization": "",
      "visual_success_criteria": ""
    }
  ],
  "creative_plan": {
    "scene": {
      "description": "",
      "visual_success_criteria": ""
      "why_fit_product": ""
      "negative_scene": ""
    },
    "style": {
      "description": "",
      "visual_success_criteria": ""
      "why_fit_product": ""
      "negative_style": ""
    },
    "shots": [
      {
        "shot_id": 1,
        "time": "",
        "shot_type": "",
        "motion_type": "",
        "description": "",
        "intent": "",
        "visual_success_criteria": "",
        "selling_point_indices": []
      }
    ]
  },
  "audio_script": [
    {
      "shot_id": 1,
      "time": "",
      "background_sound": "",
      "narration": ""
    }
  ],
  "generation_prompts": {
    "video_generation_prompt": "",
    "audio_generation_prompt": ""
  }
}
\end{lstlisting}

\end{tcolorbox}
\caption{Unified structured annotation schema used for both real-world and synthetic samples in AdSpark-300K.}
\label{sup:unified_annotation_schema}
\end{figure*}


%% file: Prompt/synthetic_planning_prompt.tex
\begin{figure*}[!t]
\centering
\begin{tcolorbox}[
    colback=gray!5,
    colframe=gray!60,
    title={\textbf{Prompt for Synthetic Advertisement Planning}},
    fonttitle=\bfseries,
    width=\textwidth,
    boxrule=0.5pt
]
\small

You are an experienced e-commerce advertisement director and video prompt engineer. A product reference image and associated metadata, including the product title, brand, category, attributes, and merchant-provided selling-point descriptions, are provided.

\textbf{Task}:  
Generate structured advertisement annotations and executable visual and audio prompts for a product-centric advertisement video. All content must be grounded in the provided product reference and metadata. Do not introduce unsupported functions, numerical claims, product properties, or commercial benefits.

\textbf{Product Identity}:  
Describe the product category, brand, product name, and identity-defining visual attributes, including dominant colors, materials, shape, key regions, recognizable logos or packaging text, and essential structural details that should remain unchanged throughout the video.

\textbf{Selling Points}:  
Use merchant-provided selling-point descriptions as the primary basis for advertisement planning. Select one to five core selling points and translate each into a concrete visual realization and an observable success criterion. Selling points must be demonstrated through visible actions, interactions, material details, or product effects rather than subtitles or narration alone.

\textbf{Creative Plan}:  
Design one to five temporally ordered shots covering the complete video duration. For each shot, specify its temporal range, shot type, camera motion, description, intent, corresponding selling points, and visual success criterion. The \texttt{shot\_type} and \texttt{motion\_type} fields must be selected from the predefined vocabularies. Shot intervals must be consecutive and non-overlapping, and the scene and visual style should be appropriate for the advertised product.

\textbf{Scene and Style}:  
Design a product-specific environment and visual style. Clearly specify the setting, props, composition, lighting, color palette, contrast, texture, and overall commercial tone. Avoid generic descriptions and ensure that the scene and style are appropriate for the advertised product.

\textbf{Video Prompt}:  
Expand the creative plan into an executable shot-by-shot prompt. For each shot, describe the framing, composition, viewpoint, focal emphasis, depth of field, camera trajectory, motion rhythm, lighting, product interaction, and visual realization of the corresponding selling points. The number, order, and timing of shots must exactly match the structured plan. All actions must follow basic physical constraints, including gravity, object support, contact relations, liquid behavior, and human--product spatial relationships.

The generated video may contain only text printed on the product packaging. Do not design subtitles, floating captions, slogans, promotional overlays, prices, interface elements, watermarks, or other non-product text.

\textbf{Audio Script}:  
Provide one audio segment for each planned shot. Each segment should contain an optional brief environmental or action-related sound and one concise narration sentence aligned with the selling point and visual content of the same shot. Do not introduce unsupported claims or use narration to replace visual demonstration.

\textbf{Cross-Field Consistency}:  
The product identity, selling points, creative plan, video-generation prompt, and audio script must describe the same advertisement. All shots, selling-point references, temporal ranges, camera motions, and audio segments must remain consistent across fields. Do not add scenes, actions, or selling points that are absent from the structured annotations.

\textbf{Output Format}:  
Return only a valid JSON object following the schema below:

\begin{lstlisting}
Return only a valid JSON object following the schema shown in Fig. 14.
\end{lstlisting}

\textbf{Input Fields}:

Product Reference Image:

Product Metadata:

Target Video Duration:

\end{tcolorbox}
\caption{\textbf{Prompt for synthetic advertisement planning with GPT-5.5.} The production prompt is condensed to retain its core product-grounding, selling-point visualization, shot-planning, and audio-alignment instructions.}
\label{sup:synthetic_planning_prompt}
\end{figure*}

%% file: Prompt/shot_camera_motion.tex
\begin{figure*}[!t]
\centering
\begin{tcolorbox}[
    colback=gray!5,
    colframe=gray!60,
    title=\textbf{Controlled Vocabulary for Shot and Motion Planning},
    fonttitle=\bfseries,
    width=\textwidth,
    boxrule=0.5pt
]
\small

The \texttt{shot\_type} and \texttt{motion\_type} fields in the synthetic
advertisement plan must be selected from the following predefined
vocabularies. The selected types should remain consistent with the shot
description, intended selling point, and executable video prompt.

\textbf{Shot Types}:

\texttt{hero\_shot}: Establishes a strong product-centered advertising visual using centered or rule-of-thirds composition.
    
\texttt{full\_product\_shot}: Presents the complete product shape, proportions, silhouette, and primary packaging face.
    
\texttt{medium\_product\_shot}: Balances the product with part of the surrounding environment, suitable for light action or everyday display.

\texttt{macro\_detail\_shot}: Highlights local product details, materials, textures, openings, applicators, liquids, sprays, or other fine-grained regions.

\texttt{demonstration\_shot}: Demonstrates product functions or usage through visible actions such as applying, pouring, spraying, opening, holding, or operating.

\texttt{interaction\_shot}: Emphasizes natural human--product interactions, including picking up, touching, opening, placing, or handing over the product.

\texttt{establishing\_shot}: Introduces the product environment and surrounding spatial context, such as a bathroom, kitchen, office, tabletop, or outdoor setting.

\texttt{lifestyle\_shot}: Places the product within a category-appropriate lifestyle scenario, such as morning care, post-workout use, home relaxation, or daily grooming.

\texttt{comparison\_shot}: Uses a visible contrast to present a product effect, such as before--after, wet--dry, clean--unclean, or texture differences.

\texttt{material\_texture\_shot}: Emphasizes surface appearance, reflections, glass, metal, plastic, fabric, liquid, cream, or particle texture.

\texttt{packshot}: Provides a stable final product display with a clean and clearly composed closing frame.

\textbf{Motion Types}:

\texttt{locked\_off}: Uses a fixed camera for stable product presentation, detail emphasis, or final framing.

\texttt{dolly\_in}: Slowly moves toward the product to emphasize the subject, selling point, or emotional progression.

\texttt{dolly\_out}: Gradually moves away to reveal the relationship between the product and its environment.

\texttt{pan\_left} / \texttt{pan\_right}: Rotates the camera horizontally to reveal the surrounding space or transition between the product and a usage action.

\texttt{tilt\_up} / \texttt{tilt\_down}: Rotates the camera vertically to reveal product height, packaging details, tabletop actions, or usage processes.

\texttt{truck\_left} / \texttt{truck\_right}: Moves the camera laterally to create spatial parallax and reveal product sides or scene context.

\texttt{orbit\_left} / \texttt{orbit\_right}: Performs a small controlled orbit around the product to reveal its three-dimensional form without a full rotation.

\texttt{tracking\_shot}: Follows the product or hand movement during a usage demonstration or human--product interaction.

\texttt{rack\_focus}: Transfers focus between the product, surrounding elements, hand actions, or fine-grained product details.

\texttt{subtle\_handheld}: Adds restrained handheld motion to create a natural documentary feel while keeping the product stable and clear.

\texttt{zoom\_in} / \texttt{zoom\_out}: Adjusts focal length to emphasize a detail or reveal the surrounding scene, while avoiding abrupt digital-zoom effects.

\textbf{Selection Rule}:  
Each planned shot must contain exactly one \texttt{shot\_type} and one
\texttt{motion\_type}. The selected pair must be compatible with the shot
description and visual objective. Do not introduce undefined shot or motion
types.

\end{tcolorbox}
\caption{Controlled vocabulary for shot and camera-motion planning in
synthetic advertisement construction.}
\label{sup:shot_motion_vocabulary}
\end{figure*}

%% file: Prompt/synthetic_plan_review.tex
\begin{figure*}[!t]
\centering
\begin{tcolorbox}[
    colback=gray!5,
    colframe=gray!60,
    title=\textbf{Prompt for Synthetic Advertisement Plan Review},
    fonttitle=\bfseries,
    width=\textwidth,
    boxrule=0.5pt
]
\small

You are an expert reviewer of product-centric advertisement plans. A product
reference image, associated product metadata, and a structured advertisement
plan intended for video generation are provided.

\textbf{Task}:  
Evaluate the advertisement plan itself before video generation. Determine
whether it accurately describes the product, uses supported selling points,
provides a coherent and executable creative plan, and remains internally
consistent across all fields. Do not evaluate or imagine the quality of the
final generated video.

\textbf{Product Identity Accuracy}:  
Verify that the product category, brand, product name, dominant colors,
materials, shape, packaging layout, logos, text, key regions, and other
identity-defining details are consistent with the provided product reference
and metadata. Reject plans that alter or invent the product identity.

\textbf{Selling-Point Validity}:  
Verify that all selected selling points are supported by the
merchant-provided descriptions, product metadata, visible product
characteristics, or reasonable category knowledge. Reject unsupported
functions, numerical claims, product properties, or commercial benefits.
Each selling point should be translated into a concrete visual realization
rather than expressed only through narration.

\textbf{Creative-Plan Coherence}:  
Verify that the planned scene, visual style, shot types, camera motions,
product interactions, and selling-point progression are appropriate for the
product and mutually consistent. The number of shots and selling points must
remain within the predefined limits. Shot intervals must be temporally
ordered, consecutive, non-overlapping, and jointly cover the complete video
duration.

\textbf{Shot and Motion Constraints}:  
Verify that every shot contains exactly one valid \texttt{shot\_type} and
one valid \texttt{motion\_type} selected from the predefined controlled
vocabularies. The selected types must agree with the shot description,
camera behavior, and intended visual objective. Reject undefined,
contradictory, or newly invented shot and motion labels.

\textbf{Physical Plausibility}:  
Verify that all described actions follow basic physical constraints,
including gravity, object support, contact relations, liquid or particle
behavior, and human--product spatial relationships. Reject floating
objects, penetration, unsupported suspension, impossible motions,
contradictory object states, or implausible interactions.

\textbf{Schema Completeness and Consistency}:  
Verify that all required fields in the unified annotation schema are
present and correctly formatted. Each shot must specify its temporal range,
shot type, motion type, description, intent, associated selling points, and
visual success criterion. Selling-point indices must refer to valid entries,
and every selling point should be covered by at least one shot. The
structured annotations, video-generation prompt, and audio-generation
prompt must describe the same plan without omissions, reordering, or
unsupported additions.

\end{tcolorbox}

\caption{\textbf{Prompt for synthetic advertisement plan review with
Gemini-3.1-Pro-Preview.}}
\label{sup:synthetic_review_prompt}
\end{figure*}

\begin{figure*}[!t]
\ContinuedFloat
\centering
\begin{tcolorbox}[
    colback=gray!5,
    colframe=gray!60,
    title=\textbf{Prompt for Synthetic Advertisement Plan Review (Continued)},
    fonttitle=\bfseries,
    width=\textwidth,
    boxrule=0.5pt
]
\small

\textbf{Visual Executability}:  
Verify that each shot provides concrete and executable descriptions of
framing, composition, viewpoint, focal emphasis, depth of field, camera
trajectory, motion rhythm, lighting, product action, and selling-point
presentation. Reject vague instructions, generic advertising language,
repetitive descriptions, or mutually contradictory visual requirements.
Each visual success criterion must describe observable evidence that can be
assessed from the generated video.

\textbf{Scene and Style Suitability}:  
Verify that the environment, props, lighting, color palette, contrast,
texture, and overall visual tone are appropriate for the product category
and intended advertising context. Reject scenes or styles that conflict
with the product attributes or introduce unsupported usage scenarios.

\textbf{Video-Prompt Alignment}:  
Verify that the video-generation prompt contains the same number, order,
types, motions, and temporal ranges of shots as the structured creative
plan. Each shot should explicitly realize its associated selling points and
success criteria. The prompt must not introduce additional shots, actions,
selling points, or scenes absent from the structured annotations.

\textbf{Audio Alignment}:  
Verify that the audio-generation prompt contains one segment for each
planned shot and follows the same shot order and temporal ranges. Each
segment should include at most one concise narration sentence and an
optional relevant sound effect. Narration and sound effects must correspond
to the visual content and selling point of the same shot and must not
introduce unsupported claims.

\textbf{Text Compliance}:  
Verify that the planned video does not require subtitles, floating captions,
promotional slogans, prices, interface elements, watermarks, or any other
text outside the product packaging.

\textbf{Decision Rule}:  
Return \texttt{yes} only when the plan is accurate, complete, physically
plausible, internally consistent, and directly executable. Otherwise,
return \texttt{no} and provide specific, actionable revision instructions.
Identify the exact fields or statements that should be corrected, removed,
or rewritten.

\textbf{Output Format}:  
Return only a valid JSON object following the schema below:

\begin{lstlisting}[
    basicstyle=\ttfamily\scriptsize,
    columns=fullflexible,
    keepspaces=true,
    breaklines=true,
    breakatwhitespace=true,
    lineskip=-1pt,
    aboveskip=2pt,
    belowskip=2pt,
    showstringspaces=false
]
{
    "verdict": "yes",
    "score": 9,
    "issues": [],
    "feedback": ""
}
\end{lstlisting}

When the plan is rejected, list the identified issues and provide concrete
revision instructions in \texttt{feedback}.

\textbf{Input Fields}:

Product Reference Image:

Product Metadata:

Structured Advertisement Annotation JSON:

\end{tcolorbox}

\caption[]{\textbf{Prompt for synthetic advertisement plan review with
Gemini-3.1-Pro-Preview (continued).}}
\end{figure*}

%% file: Prompt/ann_example_adspark_300k.tex

\begin{figure*}[!t]
\centering
\begin{tcolorbox}[
    colback=gray!5,
    colframe=gray!60,
    title={\textbf{An Annotation Example from AdSpark-300K}},
    fonttitle=\bfseries,
    width=\textwidth,
    boxrule=0.5pt,
    left=3pt,
    right=3pt,
    top=2pt,
    bottom=2pt
]

\begin{lstlisting}[
    basicstyle=\ttfamily\scriptsize,
    columns=fullflexible,
    keepspaces=true,
    breaklines=true,
    breakatwhitespace=true,
    lineskip=-1pt,
    aboveskip=2pt,
    belowskip=2pt,
    showstringspaces=false
]
{
  "product_identity": {
    "category": "Men's Mechanical Diving Watch",
    "brand": "TITONI",
    "product_name": "Seascoper Automatic Diving Watch",
    "visual_identity": {
      "main_colors": [
        "deep ocean blue",
        "silver",
        "black",
      ],
      "materials": [
        "metal case",
        "serrated rotating bezel",
        "blue dial",
        "black rubber strap",
      ],
      "shape": "A round diving-watch case with a wide black rubber strap, a prominent crown on the right, and a robust frontal profile.",
      "key_regions": [
        "triangular luminous marker at 12 o'clock",
        "brand logo at the center of the dial",
        "date window at 3 o'clock",
        "serrated rotating bezel",
        "vertical grooves on the black rubber strap"
      ],
      "ocr_targets": [
        "TITONI",
        "SEASCOPER",
        "CHRONOMETER",
        "SWISS MADE",
        "28"
      ],
      "must_keep": [
        "deep-blue dial and blue outer bezel",
        "silver case and right-side crown",
        "black rubber strap",
        "white circular and bar-shaped hour markers",
        "red detail on the second hand"
      ]
    }
  },
  "selling_points": [
    {
      "point": "Swiss-made mechanical watch",
      "visualization": "Use a stable hero shot, precise metallic highlights, clear dial layering, and finely rendered markers to convey Swiss mechanical craftsmanship.",
      "visual_success_criteria": "The complete front of the watch, refined reflections on the metal case, and orderly markers and hands should remain clearly visible, presenting it as a premium mechanical watch."
    },
    {
      "point": "Swiss chronometer certification",
      "visualization": "Use a macro shot and rack focus across the hands, date window, and certification area to emphasize precise timekeeping and certified craftsmanship.",
      "visual_success_criteria": "The video should include a close-up of the dial, with the certification area, hands, markers, or date window rendered in sharp focus."
    },
\end{lstlisting}

\end{tcolorbox}

\caption{\textbf{An annotation example from AdSpark-300K.}
The structured annotation specifies product identity, selling points,
scene and style designs, shot-level creative plans, aligned audio
scripts, and generation prompts.}
\label{sup:annotation_example_adspark_300k}
\end{figure*}

\begin{figure*}[!t]
\ContinuedFloat
\centering
\begin{tcolorbox}[
    colback=gray!5,
    colframe=gray!60,
    title={\textbf{An Annotation Example from AdSpark-300K (Continued)}},
    fonttitle=\bfseries,
    width=\textwidth,
    boxrule=0.5pt,
    left=3pt,
    right=3pt,
    top=2pt,
    bottom=2pt
]

\begin{lstlisting}[
    basicstyle=\ttfamily\scriptsize,
    columns=fullflexible,
    keepspaces=true,
    breaklines=true,
    breakatwhitespace=true,
    lineskip=-1pt,
    aboveskip=2pt,
    belowskip=2pt,
    showstringspaces=false
]
  {
      "point": "300-meter diving capability",
      "visualization": "Show the watch being submerged in clear water, with flowing water, bubbles, and underwater refraction surrounding the case while keeping the bezel and dial visible.",
      "visual_success_criteria": "The watch should visibly enter and remain in contact with water, with droplets or bubbles covering the case while the dial remains recognizable."
    }
  ],
  "creative_plan": {
    "scene": {
      "description": "A dark, wet rock surface or black matte platform beside a shallow pool of clear water, with softly blurred aquatic reflections and cool rim lighting in the background.",
      "visual_success_criteria": "The watch should maintain physically plausible support or wrist contact, while the water and platform interact naturally and the background remains visually unobtrusive.",
      "why_fit_product": "Wet rock, water droplets, and clear water naturally reinforce the diving-watch identity without interfering with product recognition.",
      "negative_scene": "Avoid cluttered office desks, promotional gift boxes, excessive text boards, floating watches, and unrelated jewelry props."
    },
    "style": {
      "description": "A refined watch-advertising style dominated by deep blue, cool silver, and black, with a restrained dark background, narrow highlights, reflective surfaces, and subtle aquatic light patterns.",
      "visual_success_criteria": "The imagery should maintain a cool blue-black palette, with thin highlights along the metal case and water reflections that do not obscure the dial.",
      "why_fit_product": "The deep-blue dial and diving functionality suit a cool ocean-inspired palette, while precise metallic highlights emphasize premium watchmaking craftsmanship.",
      "negative_style": "Avoid warm cosmetic aesthetics, exaggerated neon cyberpunk effects, cartoon rendering, aggressive promotional graphics, and cluttered high-saturation backgrounds."
    },
    "shots": [
      {
        "shot_id": 1,
        "time": "0-4 seconds",
        "shot_type": "hero shot",
        "motion_type": "dolly in",
        "description": "A low-angle hero shot establishes the complete watch standing upright on a dark, wet rock support. The deep-blue dial faces the camera while soft reflections from the water appear in the background.",
        "intent": "Establish the product as a premium men's mechanical diving watch and convey the precision and solidity associated with Swiss watchmaking.",
        "visual_success_criteria": "The case, strap, crown, serrated bezel, and blue dial should remain fully visible, with the watch standing stably and cool highlights defining the metal edges.",
        "selling_point_indices": [
          0
        ]
      },
      {
        "shot_id": 2,
        "time": "4-9 seconds",
        "shot_type": "macro detail shot",
        "motion_type": "rack focus",
        "description": "A macro shot moves from the bezel markings and hands toward the certification text near the lower dial, using shallow depth of field to emphasize the markers, date window, hands, and printed details.",
        "intent": "Present the chronometer certification and precise timekeeping through visible dial details without relying on additional on-screen text.",
        "visual_success_criteria": "The dial markers should remain orderly, the hand edges sharp, the date window clear, and the certification area positioned within the focal plane or at the endpoint of the focus transition.",
        "selling_point_indices": [1]
      },
\end{lstlisting}

\end{tcolorbox}

\caption[]{\textbf{An annotation example from AdSpark-300K
(continued).}}
\end{figure*}

\begin{figure*}[!t]
\ContinuedFloat
\centering
\begin{tcolorbox}[
    colback=gray!5,
    colframe=gray!60,
    title={\textbf{An Annotation Example from AdSpark-300K (Continued)}},
    fonttitle=\bfseries,
    width=\textwidth,
    boxrule=0.5pt,
    left=3pt,
    right=3pt,
    top=2pt,
    bottom=2pt
]

\begin{lstlisting}[
    basicstyle=\ttfamily\scriptsize,
    columns=fullflexible,
    keepspaces=true,
    breaklines=true,
    breakatwhitespace=true,
    lineskip=-1pt,
    aboveskip=2pt,
    belowskip=2pt,
    showstringspaces=false
]
      {
        "shot_id": 3,
        "time": "9-15 seconds",
        "shot_type": "demonstration shot",
        "motion_type": "tracking shot",
        "description": "A wrist wearing the watch slowly enters a clear shallow pool. The camera follows the wrist as it descends, while flowing water, bubbles, and refracted light move across the bezel and case.",
        "intent": "Demonstrate the watch's diving functionality and suitability for aquatic exploration through direct and physically plausible water interaction.",
        "visual_success_criteria": "The watch should remain securely attached to the wrist and fully submerged, with visible droplets, bubbles, and refraction while the dial remains clear and free from deformation.",
        "selling_point_indices": [
          2
        ]
      }
    ]
  },

  "audio_script": [
    {
      "shot_id": 1,
      "time": "0-4 seconds",
      "background_sound": "A deep metallic resonance.",
      "narration": "Swiss mechanical precision, built to make a bold entrance."
    },
    {
      "shot_id": 2,
      "time": "4-9 seconds",
      "background_sound": "Subtle mechanical ticking and gear sounds.",
      "narration": "Chronometer-certified precision you can trust."
    },
    {
      "shot_id": 3,
      "time": "9-15 seconds",
      "background_sound": "The sound of a wrist entering clear water.",
      "narration": "Water-resistant to 300 meters, ready to explore the deep."
    }
  ],

  "generation_prompts": {
    "video_generation_prompt": "Create a premium diving-watch advertisement in a deep-ocean-blue, cool-silver, and black visual style. The setting consists of a wet black rock surface, a matte platform, and a shallow pool of clear water, with soft aquatic reflections in the blurred background and narrow rim lighting defining the metal edges. Shot 1, hero shot with dolly-in from 0 to 4 seconds: show the complete watch standing upright on the wet rock support in a low-angle medium close-up. Keep the deep-blue dial facing the camera, the black rubber strap extending downward, and the focus locked on the dial and bezel. Slowly push the camera forward while cool highlights move across the silver case and serrated bezel. Shot 2, macro detail shot with rack focus from 4 to 9 seconds: begin with the white hour markers, dimensional hands, and date-window frame in sharp focus, then smoothly shift focus toward the certification area and minute markings near the lower dial. Add a soft highlight across the glass without obscuring the details. Shot 3, demonstration shot with tracking motion from 9 to 15 seconds: show a wrist wearing the watch slowly entering the clear shallow pool from the right side of the frame. Follow the wrist steadily as it descends, with water droplets, small bubbles, and flowing refraction moving across the bezel and case. End with the wrist resting beneath the water while the blue dial and silver case remain clearly recognizable.",
    "audio_generation_prompt": "Shot 1, 0-4 seconds: deep metallic resonance. Narration: 'Swiss mechanical precision, built to make a bold entrance.' Shot 2, 4-9 seconds: subtle mechanical ticking and gear sounds. Narration: 'Chronometer-certified precision you can trust.' Shot 3, 9-15 seconds: clear water-entry sound. Narration: 'Water-resistant to 300 meters, ready to explore the deep.'"
  }
}
\end{lstlisting}

\end{tcolorbox}

\caption[]{\textbf{An annotation example from AdSpark-300K
(continued).}}
\end{figure*}

%% file: Prompt/ann_example_adspark-bench.tex

\begin{figure*}[!t]
\centering
\begin{tcolorbox}[
    colback=gray!5,
    colframe=gray!60,
    title={\textbf{An Annotation Example from AdSpark-Bench}},
    fonttitle=\bfseries,
    width=\textwidth,
    boxrule=0.5pt,
    left=3pt,
    right=3pt,
    top=2pt,
    bottom=2pt
]

\begin{lstlisting}[
    basicstyle=\ttfamily\scriptsize,
    columns=fullflexible,
    keepspaces=true,
    breaklines=true,
    breakatwhitespace=true,
    lineskip=-1pt,
    aboveskip=2pt,
    belowskip=2pt,
    showstringspaces=false
]
{
  "product_identity": {
    "category": "Mirrorless Camera Body",
    "brand": "OLYMPUS",
    "product_name": "OM-D E-M10 Mark IV Black Body",
    "visual_identity": {
      "main_colors": [
        "black",
        "silver metallic",
        "deep-purple sensor reflection"
      ],
      "materials": [
        "matte black camera body",
        "leather-textured grip",
        "silver metal lens mount",
        "reflective glass sensor surface"
      ],
      "shape": "A compact mirrorless camera body with a retro SLR-style silhouette, a raised central viewfinder housing, and a circular lens mount centered on the front.",
      "key_regions": [
        "OLYMPUS logo on the top housing",
        "OM-D logo in the upper-left area",
        "central silver lens mount",
        "purple sensor window",
        "IV badge in the lower-right area",
        "dual control dials on the top"
      ],
      "ocr_targets": [
        "OLYMPUS",
        "OM-D",
        "IV"
      ],
      "must_keep": [
        "proportions between the black body and silver mount",
        "front-facing body without an attached lens",
        "top control dials and retro viewfinder lines",
        "leather-textured grip"
      ]
    }
  },

  "selling_points": [
    {
      "point": "Stylish retro appearance",
      "visualization": "Use a centered low-angle dolly-in while soft side lighting moves across the control dials, leather-textured grip, and silver lens mount against a blurred photography workbench.",
      "visual_success_criteria": "The retro viewfinder silhouette, control dials, leather texture, and metallic lens mount should remain clearly visible within a restrained and refined color palette."
    },
    {
      "point": "Stable handheld shooting with in-body stabilization",
      "visualization": "Show a hand lifting and moving the camera body over a short distance while the camera remains sharp and the background exhibits slight relative motion and blur.",
      "visual_success_criteria": "The hand should maintain natural contact with the camera, the body should remain stable and recognizable during motion, and the background should show subtle displacement or motion blur."
    }
  ],
\end{lstlisting}

\end{tcolorbox}

\caption{\textbf{An annotation example from AdSpark-Bench.}
The structured annotation specifies product identity, selling points,
scene and style designs, shot-level creative plans, aligned audio
scripts, and generation prompts.}
\label{sup:annotation_example_adspark_bench}
\end{figure*}

\begin{figure*}[!t]
\ContinuedFloat
\centering
\begin{tcolorbox}[
    colback=gray!5,
    colframe=gray!60,
    title={\textbf{An Annotation Example from AdSpark-Bench (Continued)}},
    fonttitle=\bfseries,
    width=\textwidth,
    boxrule=0.5pt,
    left=3pt,
    right=3pt,
    top=2pt,
    bottom=2pt
]

\begin{lstlisting}[
    basicstyle=\ttfamily\scriptsize,
    columns=fullflexible,
    keepspaces=true,
    breaklines=true,
    breakatwhitespace=true,
    lineskip=-1pt,
    aboveskip=2pt,
    belowskip=2pt,
    showstringspaces=false
]
  "creative_plan": {
    "scene": {
      "description": "A wooden photography workbench illuminated by soft window light, with a small number of film rolls and a camera strap appearing as blurred background props.",
      "visual_success_criteria": "The camera body should rest stably on the wooden table or be naturally supported by a hand. Background props should remain sparse and blurred, with spatial relationships consistent with gravity and physical contact.",
      "why_fit_product": "The workbench, film rolls, and camera strap reinforce the photographic and retro character of the product while keeping attention on the camera body.",
      "negative_scene": "Avoid crowded retail environments, unrelated kitchens or bathrooms, excessive promotional signs, screen interfaces, and unsupported floating products."
    },
    "style": {
      "description": "A refined retro photography-equipment advertising style with a low-saturation black-and-gold palette, soft side and back lighting, and restrained contrast.",
      "visual_success_criteria": "The black leather texture, brushed reflections on the silver lens mount, and highlights along the top silhouette should remain visible without overexposure.",
      "why_fit_product": "The black retro camera body relies on mechanical details, materials, and silhouette for its visual appeal, while warm workbench lighting reinforces a collectible and lifestyle-oriented character.",
      "negative_style": "Avoid saturated neon cyberpunk aesthetics, strong lens flares that obscure logos, cartoon or plastic-toy rendering, and exaggerated whip-pan motions."
    },
    "shots": [
      {
        "shot_id": 1,
        "time": "0-2 seconds",
        "shot_type": "hero shot",
        "motion_type": "dolly in",
        "description": "A slightly low-angle frontal hero shot slowly approaches the complete camera body, emphasizing the retro viewfinder housing, top control dials, leather texture, and silver lens mount.",
        "intent": "Establish a stylish retro product image and allow viewers to immediately recognize the product as a refined black mirrorless camera body.",
        "visual_success_criteria": "The complete frontal silhouette, OLYMPUS and OM-D logos, silver lens mount, leather-textured grip, and top control dials should remain clearly visible throughout a smooth camera push-in.",
        "selling_point_indices": [
          0
        ]
      },
\end{lstlisting}

\end{tcolorbox}

\caption[]{\textbf{An annotation example from AdSpark-Bench
(continued).}}
\end{figure*}

\begin{figure*}[!t]
\ContinuedFloat
\centering
\begin{tcolorbox}[
    colback=gray!5,
    colframe=gray!60,
    title={\textbf{An Annotation Example from AdSpark-Bench (Continued)}},
    fonttitle=\bfseries,
    width=\textwidth,
    boxrule=0.5pt,
    left=3pt,
    right=3pt,
    top=2pt,
    bottom=2pt
]

\begin{lstlisting}[
    basicstyle=\ttfamily\scriptsize,
    columns=fullflexible,
    keepspaces=true,
    breaklines=true,
    breakatwhitespace=true,
    lineskip=-1pt,
    aboveskip=2pt,
    belowskip=2pt,
    showstringspaces=false
]
      {
        "shot_id": 2,
        "time": "2-5 seconds",
        "shot_type": "demonstration shot",
        "motion_type": "tracking shot",
        "description": "A hand naturally grips the camera body, lifts it from the workbench, and moves it over a short distance while the filming camera smoothly follows the motion.",
        "intent": "Translate the abstract in-body stabilization feature into a visible handheld shooting experience and transition the product from static display to realistic use.",
        "visual_success_criteria": "The hand and camera should maintain physically plausible contact. The camera body should remain sharp and stable while the wooden surface and background props show subtle relative motion or blur.",
        "selling_point_indices": [
          1
        ]
      }
    ]
  },

  "audio_script": [
    {
      "shot_id": 1,
      "time": "0-2 seconds",
      "background_sound": "A subtle click from the camera control dial.",
      "narration": "Retro character that captivates at first sight."
    },
    {
      "shot_id": 2,
      "time": "2-5 seconds",
      "background_sound": "A soft sound of the camera being gripped and lifted.",
      "narration": "Steady in your hands, effortless in every shot."
    }
  ],

  "generation_prompts": {
    "video_generation_prompt": "Create a refined advertisement for a black retro mirrorless camera body on a wooden photography workbench in soft morning window light. Use a restrained low-saturation black-and-gold palette, with a few blurred film rolls and a black camera strap in the background. Shot 1, hero shot with dolly-in from 0 to 2 seconds: place the camera body at the center of the wooden table in a slightly low-angle frontal view. Slowly push toward the product while keeping focus on the OLYMPUS logo, OM-D logo, central silver lens mount, top control dials, retro viewfinder silhouette, and leather-textured grip. Allow soft side and back lighting to define the metallic mount and body contours, with a subtle purple reflection on the exposed sensor. Shot 2, demonstration shot with tracking motion from 2 to 5 seconds: show one hand naturally gripping the camera body from both sides and gently lifting it from the table. Follow the camera smoothly as the hand moves it over a short distance. Keep the silver mount edge, body silhouette, and printed logos sharp and stable, while the wooden surface and background photography props exhibit slight relative motion and blur to convey stable handheld shooting.",
    "audio_generation_prompt": "Shot 1, 0-2 seconds: a subtle camera-dial click. Narration: 'Retro character that captivates at first sight.' Shot 2, 2-5 seconds: a soft gripping and lifting sound. Narration: 'Steady in your hands, effortless in every shot.'"
  }
}
\end{lstlisting}

\end{tcolorbox}

\caption[]{\textbf{An annotation example from AdSpark-Bench
(continued).}}
\end{figure*}

%% file: Prompt/shot_execution_alignment.tex
\begin{figure*}[!t]
\centering
\begin{tcolorbox}[
    colback=gray!5,
    colframe=gray!60,
    title={\textbf{Prompt for Shot Execution Alignment}},
    fonttitle=\bfseries,
    width=\textwidth,
    boxrule=0.5pt
]
\small

You are an expert evaluator for product advertisement video shots. The provided contact sheets contain all frames from a single generated shot, arranged in temporal order from left to right and from top to bottom. Each frame is indexed. Please inspect the temporal evolution of the shot to infer the camera motion, shot type, framing quality, and whether the shot fulfills its advertisement intent.

\textbf{Planned Shot}:  

Expected shot type: \{shot\_type\}.  

Expected motion type: \{motion\_type\}.  

Shot description: \{shot\_description\}.  

Shot intent: \{shot\_intent\}.  

Optical-flow-based motion estimate: \{flow\_motion\_type\}. This estimate is only a reference and may be inaccurate.

\textbf{Task}:  
Evaluate whether the generated shot follows the planned shot type, camera motion, composition, and intended advertisement purpose. Use the full temporal contact sheet rather than a single frame. Be strict and discriminative, and do not assign high scores merely because the image looks visually pleasant.

\textbf{Output Format}:  
Return only a valid JSON object with the following keys:

\begin{lstlisting}
{
  "shot_type_pred": "...",
  "shot_type_reason": "...",
  "shot_type_match": 0.0-1.0,
  "motion_type_pred": "...",
  "motion_type_reason": "...",
  "motion_type_match": 0.0-1.0,
  "framing_quality": 0.0-1.0,
  "intent_fulfillment": 0.0-1.0
}
\end{lstlisting}

\textbf{Input Fields}:  

Temporal Contact Sheets:

Planned Shot Type:

Planned Motion Type:

Shot Description:

Shot Intent:

Optical-flow Motion Estimate:

\end{tcolorbox}
\caption{Prompt for GPT-5.5-based shot execution alignment.}
\label{sup:shot_execution_prompt}
\end{figure*}

%% file: Prompt/scene_style_alignment.tex
\begin{figure*}[!t]
\centering
\begin{tcolorbox}[
    colback=gray!5,
    colframe=gray!60,
    title={\textbf{Prompt for Scene and Style Alignment}},
    fonttitle=\bfseries,
    width=\textwidth,
    boxrule=0.5pt
]
\small

The provided contact sheets contain representative frames sampled from each shot of a product advertisement video. The frames are arranged in temporal order. The product information, expected scene, and expected style from the structured creative plan are also provided.

\textbf{Product}:  
\{product\_identity\}

\textbf{Expected Scene}:  
\{scene\_description\}

Why this scene fits the product: \{scene\_rationale\}

Negative scene examples: \{negative\_scene\}

Visual success criteria: \{scene\_success\_criteria\}

\textbf{Expected Style}:  
\{style\_description\}

Why this style fits the product: \{style\_rationale\}

Negative style examples: \{negative\_style\}

Visual success criteria: \{style\_success\_criteria\}

\textbf{Task}:  
You are evaluating whether the generated video satisfies the scene and style requirements in the creative plan. Judge only whether the visual content follows the provided schema. Do not evaluate general visual quality or product identity preservation, since these are measured by other metrics.

For each item, first check the visual success criteria one by one. Consider whether the visual environment, lighting, color tone, composition, background, props, and atmosphere are consistent with the expected scene and style. If the generated video matches any negative example, assign a low score.

\textbf{Scoring Rule}:  
Use integer scores from 1 to 5.  

1 = almost completely inconsistent or violates negative examples.  

3 = roughly related but only partially satisfies the criteria.  

5 = strongly matches the expected scene or style with clear visual evidence.

\textbf{Output Format}:  
Return only a valid JSON object:

\begin{lstlisting}
{
  "scene_alignment_comment": "...",
  "scene_alignment": 1-5,
  "style_alignment_comment": "...",
  "style_alignment": 1-5
}
\end{lstlisting}

\textbf{Input Fields}:  

Shot-level Contact Sheets:

Product Identity:

Expected Scene:

Expected Style:

Visual Success Criteria:

\end{tcolorbox}
\caption{Prompt for GPT-5.5-based scene and style alignment.}
\label{sup:scene_style_prompt}
\end{figure*}

%% file: Prompt/selling_point_realization.tex
\begin{figure*}[!t]
\centering
\begin{tcolorbox}[
    colback=gray!5,
    colframe=gray!60,
    title={\textbf{Prompt for Selling-point Realization}},
    fonttitle=\bfseries,
    width=\textwidth,
    boxrule=0.5pt
]
\small

The provided contact sheets densely cover the generated advertisement video in temporal order. The product information and the planned selling points are also provided. Each selling point includes its intended visual presentation and observable success criteria.

\textbf{Product}:  
\{product\_identity\}

\textbf{Selling Points}:  
\{selling\_points\}

For each selling point, the input may include:

Point description: \{point\_i\}.  

Visualization plan: \{visualization\_i\}.  

Visual success criteria: \{success\_criteria\_i\}.  

\textbf{Task}:  
Evaluate whether each planned selling point is clearly presented or demonstrated anywhere in the video. A selling point does not need to appear in a specific shot; it is sufficient if it is clearly realized at any moment in the generated video. Judge only visible evidence from the frames. Do not reward unsupported product claims or invisible functions. Do not evaluate product identity preservation here.

\textbf{Scoring Rule}:  
Use integer scores from 1 to 5 for each selling point.  

1 = the selling point is missing or almost invisible.  

3 = the selling point appears but is ambiguous, incomplete, or only partially satisfies the success criteria.  

5 = the selling point is clearly and effectively visualized according to the planned criteria.

\textbf{Output Format}:  
Return only a valid JSON object. Use one key for each selling point:

\begin{lstlisting}
{
  "point_0_comment": "...",
  "point_0": 1-5,
  "point_1_comment": "...",
  "point_1": 1-5
}
\end{lstlisting}

\textbf{Input Fields}:  

Dense Shot-level Contact Sheets:

Product Identity:

Selling Points:

Visualization Plans:

Visual Success Criteria:

\end{tcolorbox}
\caption{Prompt for GPT-5.5-based selling-point realization.}
\label{sup:selling_point_prompt}
\end{figure*}

%% file: Prompt/transition_naturalness.tex
\begin{figure*}[!t]
\centering
\begin{tcolorbox}[
    colback=gray!5,
    colframe=gray!60,
    title={\textbf{Prompt for Transition Naturalness}},
    fonttitle=\bfseries,
    width=\textwidth,
    boxrule=0.5pt
]
\small

You are evaluating a shot transition in a product advertisement video. The provided images contain two frames before the cut and two frames after the cut.

\textbf{Shot Context}:  

Shot before cut description: \{previous\_shot\_description\}.  

Shot after cut description: \{next\_shot\_description\}.

\textbf{Task}:  
Rate how natural the transition is. Consider brightness and color consistency, absence of corrupted pixels, stylistic coherence, and whether the cut appears intentional. A valid scene change should not be penalized. Penalize only visual artifacts, flickering, corrupted frames, or jarring style breaks.

\textbf{Output Format}:  
Reply with a single float from 0.0 to 1.0, where 1.0 means perfectly natural. Output only the number, without any additional text.

\textbf{Input Fields}:  

Two Before-cut Frames:

Two After-cut Frames:

Previous Shot Description:

Next Shot Description:

\end{tcolorbox}
\caption{Prompt for GPT-5.5-based transition naturalness evaluation.}
\label{sup:transition_prompt}
\end{figure*}

%% file: Prompt/advertisement_attractive.tex
\begin{figure*}[!t]
\centering
\begin{tcolorbox}[
    colback=gray!5,
    colframe=gray!60,
    title={\textbf{Prompt for Advertisement Attractiveness}},
    fonttitle=\bfseries,
    width=\textwidth,
    boxrule=0.5pt
]
\small

The provided frames are sampled from a product advertisement video in temporal order. The product information is also provided. Please evaluate the video from the perspective of an ordinary customer who encounters this advertisement in an e-commerce or short-video feed.

\textbf{Product}:  
\{product\_identity\}

\textbf{Task}:  
Evaluate the attractiveness of the generated advertisement as a commercial video. Ignore whether the video exactly follows the planned creative script; focus on how appealing and engaging it is to a potential customer.

\textbf{Evaluation Items}:  

\textbf{visual\_appeal}: Whether the video has commercial advertisement-level visual appeal rather than a cheap or casual generated look.  

\textbf{attention\_capture}: Whether the video has a clear visual focus and dynamic changes that can attract attention in the first few seconds.  

\textbf{product\_desirability}: Whether the product is presented as professional, clean, trustworthy, and desirable.  

\textbf{watch\_completion}: Whether the video has enough progression and rhythm to encourage viewers to watch it to completion.

\textbf{Scoring Rule}:  
Use integer scores from 1 to 5.  

1 = poor or almost missing.  

3 = acceptable but ordinary or only partially effective.  

5 = highly attractive and commercially effective.

\textbf{Output Format}:  
Return only a valid JSON object:

\begin{lstlisting}
{
  "visual_appeal_comment": "...",
  "visual_appeal": 1-5,
  "attention_capture_comment": "...",
  "attention_capture": 1-5,
  "product_desirability_comment": "...",
  "product_desirability": 1-5,
  "watch_completion_comment": "...",
  "watch_completion": 1-5
}
\end{lstlisting}

\textbf{Input Fields}:  

Temporally Ordered Sampled Frames:

Product Identity:

\end{tcolorbox}
\caption{Prompt for GPT-5.5-based advertisement attractiveness evaluation.}
\label{sup:attractiveness_prompt}
\end{figure*}

%% file: Prompt/creative_quality.tex
\begin{figure*}[!t]
\centering
\begin{tcolorbox}[
    colback=gray!5,
    colframe=gray!60,
    title={\textbf{Prompt for Creative Quality}},
    fonttitle=\bfseries,
    width=\textwidth,
    boxrule=0.5pt
]
\small

The provided frames are sampled from a product advertisement video in temporal order. The product information is also provided. Please evaluate whether the generated video has polished, commercially usable advertisement design.

\textbf{Product}:  
\{product\_identity\}

\textbf{Task}:  
Evaluate the creative and commercial quality of the advertisement. Artistic quality here refers to advertisement-grade composition, lighting, color, texture, and visual design, rather than generic prettiness.

\textbf{Evaluation Items}:  

\textbf{artistic\_quality}: Whether composition, lighting, color, and texture reach the quality expected from a commercial advertisement.  

\textbf{commercial\_readiness}: Whether the video looks like a finished advertisement that could be directly used or requires substantial revision.  

\textbf{visual\_impact}: Whether the video has memorable visual elements or dynamic impact that can stand out in a content feed.  

\textbf{product\_relevance}: Whether the creative choices, shots, and motions serve the product and its selling points rather than unrelated visual effects.

\textbf{Scoring Rule}:  
Use integer scores from 1 to 5.  

1 = poor, irrelevant, or not commercially usable.  

3 = acceptable but ordinary or requiring refinement.  

5 = polished, commercially ready, and strongly product-oriented.

\textbf{Output Format}:  
Return only a valid JSON object:

\begin{lstlisting}
{
  "artistic_quality_comment": "...",
  "artistic_quality": 1-5,
  "commercial_readiness_comment": "...",
  "commercial_readiness": 1-5,
  "visual_impact_comment": "...",
  "visual_impact": 1-5,
  "product_relevance_comment": "...",
  "product_relevance": 1-5
}
\end{lstlisting}

\textbf{Input Fields}:  

Temporally Ordered Sampled Frames:

Product Identity:

\end{tcolorbox}
\caption{Prompt for GPT-5.5-based creative quality evaluation.}
\label{sup:creative_quality_prompt}
\end{figure*}

%% file: Prompt/narr_coherence.tex
\begin{figure*}[!t]
\centering
\begin{tcolorbox}[
    colback=gray!5,
    colframe=gray!60,
    title={\textbf{Prompt for Narrative Coherence}},
    fonttitle=\bfseries,
    width=\textwidth,
    boxrule=0.5pt
]
\small

You are evaluating a multi-shot product advertisement video. The provided contact sheets tile frames in temporal order. Frame numbers are mapped to detected shots, so please use the shot grouping to understand the structure and rhythm of the video.

\textbf{Planned Shot Sequence}:  
\{planned\_shot\_sequence\}

Each planned shot may include:

Shot description: \{shot\_description\}.  

Shot intent: \{shot\_intent\}.

\textbf{Detected Shot Mapping}:  
\{frame\_to\_shot\_mapping\}

\textbf{Task}:  
Evaluate the narrative coherence of the whole generated advertisement. Consider whether the visual story flows logically from shot to shot, whether the scene and style remain consistent, whether the sequence builds toward a clear advertising message, and whether the emotional or rhythmic tone is coherent. Do not penalize a normal shot change if it supports the planned advertisement structure.

\textbf{Output Format}:  
Reply with a single float from 0.0 to 1.0, where 1.0 means a highly coherent advertisement narrative. Output only the number, without any additional text.

\textbf{Input Fields}:  

Shot-aware Contact Sheets:

Planned Shot Sequence:

Detected Shot Mapping:

Shot Descriptions:

Shot Intents:

\end{tcolorbox}
\caption{Prompt for GPT-5.5-based narrative coherence evaluation.}
\label{sup:narrative_prompt}
\end{figure*}